\documentclass[nonblindrev]{informs3mod}

\OneAndAHalfSpacedXI

\usepackage{subfig, xspace,booktabs,multirow,enumitem,appendix}

\usepackage{natbib}
 \bibpunct[, ]{(}{)}{,}{a}{}{,}
 \def\bibfont{\small}

\TheoremsNumberedThrough     
\ECRepeatTheorems

\EquationsNumberedThrough    

\usepackage{shortcuts}

\newcommand{\ba}{\begin{array}}
\newcommand{\ea}{\end{array}}
\newcommand{\bbm}{\begin{bmatrix}}
\newcommand{\ebm}{\end{bmatrix}}

\usepackage{algorithm}
\usepackage{algorithmicx}
\usepackage{algpseudocode}

\usepackage[dvipsnames]{xcolor}
\newcommand{\edit}{}
\newcommand{\bedit}{}
\newcommand{\eedit}{}
\newcommand{\editb}{}
\newcommand{\beditb}{}
\newcommand{\eeditb}{}

\begin{document}

\RUNAUTHOR{Kallus and Udell}

\RUNTITLE{Dynamic Assortment Personalization in High Dimensions}

\TITLE{Dynamic Assortment Personalization\\in High Dimensions}

\ARTICLEAUTHORS{
\AUTHOR{Nathan Kallus}
\AFF{Cornell University and Cornell Tech, \EMAIL{kallus@cornell.edu}}
\AUTHOR{Madeleine Udell}
\AFF{Cornell University, \EMAIL{udell@cornell.edu}}
}

\ABSTRACT{
We study the problem of dynamic assortment personalization with large, heterogeneous populations and wide arrays of products, and demonstrate the importance of structural priors for effective, efficient large-scale personalization. Assortment personalization is the problem of choosing, for each {individual (type),} a best assortment of products, ads, or other offerings (items) so as to maximize revenue. This problem is central to revenue management in {e-commerce and online advertising
where both items and types can number in the millions.}

We formulate the dynamic assortment personalization problem as a discrete-contextual bandit with $m$ contexts {(types)} and exponentially many arms (assortments of the $n$ items). We assume that each type's preferences follow a simple parametric model with $n$ parameters. In all, there are $mn$ parameters, and existing literature suggests that order optimal regret scales as $mn$. However, the data required to estimate so many parameters is orders of magnitude larger than the data available in most revenue management applications; and the optimal regret under these models is unacceptably high.

In this paper, we impose a natural structure on the problem -- a small latent dimension, or low rank. In the static setting, we show that this model can be efficiently learned from surprisingly few interactions, using a time- and memory-efficient optimization algorithm that converges globally whenever the model is learnable. In the dynamic setting, we show that structure-aware dynamic assortment personalization can have regret that is an order of magnitude smaller than structure-ignorant approaches. We validate our theoretical results empirically.
}


\maketitle

\section{Introduction}\label{introsec}

In many commerce, e-commerce, and advertising settings, customers or users
are presented with an assortment of products, ads, or other offerings.
Customers choose which to products to buy, ads to click, or (generically)
items to interact with, from among the assortment that is presented.
Firms choose which assortment to present to the customer, and collect the revenue
(or other benefit or loss) resulting from each customer's choice.
Choosing the assortment that maximizes expected revenue
is a central problem in revenue management.
This problem goes by the name \emph{assortment planning} or \emph{assortment optimization}.
When assortments are tailored to each individual customer or
to each consumer
segment, the problem is known as \emph{assortment personalization}.
Successful personalization, which is key
to e-commerce and online advertising operations,
hinges on learning the preferences of
each customer for each offering.

This paper shows how to learn customer preferences and manage revenue
under realistic assumptions about the problem data available
and the structure of customer preferences.
We suppose that we must rely on \emph{transactional data} to estimate
customer preferences; and that we rely only on the \emph{discrete context} of each
customer's type {(usually, each customer's unique id)}
to infer their preferences, for we lack covariates that
precisely predict customer outcomes.
These assumptions make our problem challenging.
We show how to achieve good performance using the one key structural assumption:
that customer preferences are \emph{low rank}.
We provide both new algorithms and new results for this setting
that enable efficient assortment personalization in the face of
high dimensions.

\subsection{Problem setting}

We next explain the problem setting and discuss the
practical importance of each of the main assumptions.
The problem setting we consider in this paper is,
we believe, the relevant one for most commerce and e-commerce settings.

\paragraph{Transactional data.}
Transactional data --- which customer selected which product ---
is abundant and easy for retailers to collect,
whereas detailed information about customer attributes
(such as gender, age, ethnicity, and preferences)
cannot be directly harvested from the retailer's data.
It is natural and even expected that a data driven modern retailer will record
transactional data, whereas retailers who buy information on their customers'
attributes face extra costs as well as data quality and privacy concerns.
Moreover, these covariates are not directly related to the task at hand: predicting
which products the customer will buy, and maximizing revenue.
In particular, they may be too coarse to understand
customer preferences precisely.
{
For example, even if gender and age information is available, providing
the same personalized assortment to all women aged 30 may
not be a good strategy when preferences differ.
Instead, our approach requires no side information about the attributes of the customers.
We rely exclusively on transactional data and
focus on personalization at the finest level of segmentation
offered by the recorded data (usually, the individual level), using
past behavior rather than characteristics to predict future behavior.}

\paragraph{Discrete context}
{In this paper, we define the customer's \emph{type}
to be the smallest group that is uniquely
identifiable given the available data, or anything coarser.
This notion is easily applied to the e-commerce setting:
here, the retailer records a unique user id as part of each transaction,
so each type corresponds to a single customer.
This e-commerce setting is the focus of our paper,
and the key issue we address is that the number of both types and items may be very large.
In a less common setting also covered by our framework,
multi-location brick-and-mortar stores with very many locations,
one may consider type to be the store location itself
with the intent of aggregating data to learn geographically-varying tastes.
}
An algorithm relying exclusively on transactional data lacks any context to use to
predict a customer's behavior other than the type (\eg, customer id) of that customer.
Hence we refer to our assortment choice problem as having a \emph{discrete context} corresponding to this distinct identity.
This contrasts usual contextual variables that are continuous
vectors related to one another in terms of metric proximity.

\paragraph{Regret analysis.}
This paper presents a regret analysis of assortment choices in the
discrete-contextual setting.
In this setting, ours is the first algorithm
that can achieve regret which grows sublinearly {not only in the
horizon but also} in the number of parameters (item-type combinations).
{The regret of an algorithm is the difference between
its cumulative performance
and the performance of an idealized method which knows the individual preferences of the customer
and always takes the optimal action.
(Regret is defined formally in Sec.~\ref{mainprobsec}.)
That our regret grows sublinearly in the number of item-type combinations means
that our time-average performance reaches optimality even in the high-dimensional regime
where the number of item-type combinations is large in comparison to the horizon.
We further show that no algorithm that ignores the low rank structure of the problem
can achieve sublinear regret in the number of item-type combinations.
{
Therefore, since regret in our context amounts to a difference in revenues, the additional \emph{revenue} generated by our algorithm relative to these is \emph{linear} in the time horizon in the high dimensional regime.
}
}


\paragraph{Low-rank structure.} Our method makes use of one key structural assumption
--- that preferences are (approximately) low rank --- that has been extensively verified
in a wide variety of practical applications. These applications range
from its earliest uses, in psychology \citep{spearman1904, hotelling1933},
to modern applications in marketing \citep{funk2006},
genomics \citep{witten2009}, and
healthcare \citep{schuler2016discovering}.
Indeed, under a very natural model for preferences --- roughly, as long as customers
and products are iid (independent and identically distributed), and there exists some function mapping customer attributes and
product attributes to latent utilities --- then a table of customer preferences
will be approximately low rank \citep{udell2017nice}.
This low-rank structural assumption provides two advantages over previous approaches.

First, it undergirds our regret bound, which guarantees that the regret incurred
by our algorithm grows sublinearly in the number of parameters.
This result provides a significant advance with respect to the previous literature
in both assortment optimization and in matrix sensing.
Relative to existing results in assortment optimization, our analysis provides
a better scaling as the dimension of the problem grows.
Indeed, we prove
that any structure-ignorant method must incur regret that grows linearly in the
number of preference parameters.
Relative to existing results in matrix sensing, our analysis provides the
first regret bounds for matrix sensing under bandit feedback.
Furthermore, we show how to provably learn a preference matrix in the realistic
setting of transactional data, in which we see only which product (if any)
the customer has chosen to purchase, rather than a full preference list or a forced choice.

Second, our algorithms make use of the low rank structural assumption to
scale to extremely large data sets while preserving provable optimality guarantees.
As a result, our methods are well suited to use in practical applications,
and can easily be used by any modern retailer.
While the assumptions that underly our theoretical guarantees
cannot be verified directly,
these results demonstrate that the algorithms we propose are effective in practice.

\subsection{Our approach}
In this paper, we propose a new approach to assortment personalization.
To enable tractable estimation of a personalized model with limited data,
we propose a new structural model.
In this model, the choices of each type are governed by its own personalized preference vector
(with one dimension for every item);
but these preference vectors span (or lie close to) only a low-dimensional subspace.
We demonstrate how to estimate the parameters of this model with computationally tractable
algorithms, and provide a proof of recovery with high-probability from few samples (sublinear in number type-item combinations).
Numerically, we show that given the same data, our estimator performs much better than standard maximum likelihood estimators.

We then leverage our new model to tackle the
\emph{dynamic} assortment personalization problem:
starting with \emph{no} data, how should we choose assortments to offer to different
types to maximize total profit, or equivalently, to minimize total regret?
We show theoretically and numerically that our algorithm achieves regret orders of magnitude smaller than standard methods based
either on a single multinomial logit (MNL) model or several decoupled MNL models
for each type, which we call ``structure-ignorant'' methods.
For example, when the parameter matrix has rank $r$, we achieve regret of order $r\max(m,n)\log T$ after $T$ interactions, where ignoring structure would have yielded regret of order $\min(m,n)\max(m,n)\log T$.
{(Note that for these to make sense, we let $m$, $n$, and $T$ all vary simultaneously.)}
Our results demonstrate that assortment personalization enables orders of magnitude better performance than competing approaches, and can be achieved with a tractable, efficient estimator.

All proofs are given in the appendix.

\section{Problem statement}\label{mainprobsec}

In this section we describe the problem of \emph{dynamic assortment personalization}.
We consider a problem with $m$ types and $n$ items.
During each interaction, a consumer arrives; the retailer presents the consumer with
an assortment of no more than $K$ items;
and the consumer chooses one of the items presented or chooses nothing, which we refer to as choosing the $0\thh$ item.
{The expected revenue generated when type $i$ chooses item $j$ is $W_{ij}$ and is 0 if no item is chosen.}
A problem instance is also described by two additional parameters $\mu^\star\in\Delta^m=\braces{\alpha\in\R m_+:\sum_{i=1}^m\alpha_i=1}$ and $\Theta^\star\in\R{m\times n}$
that are not known and must be estimated from data.
The parameter $\mu^\star$ describes how often each type arrives,
while the matrix $\Theta^\star$ of preference parameters governs $m$ MNL choice models, one for each type.
The number of parameters in the model is $mn + m = \Theta(mn)$.

The problem proceeds as follows.
At first, the retailer knows only $m$, $n$, and $W\in\R{m\times n}$.
Then for each interaction $t=1,2,\dots$:
\begin{enumerate}
\item {Customer of type $i_t \in \{1,\dots,m\}$ arrives with probability $\mu^\star_i$. The retailer observes the type.}
\item The retailer chooses any subset of products $S_t\subset\{1,\dots,n\}$ with $\abs{S_t}\leq K$. 
\item The customer chooses an item $j_t$ from $\{0,1,\dots,n\}$ with probability proportional to
$$
\op{weight}(j)=\left\{\ba{lc}
1&\quad\quad{j=0}\\
0&\quad\quad{j\neq0,\,j\notin S_t}\\
\exp(\Theta^\star_{i_tj})&\quad\quad{j\neq0,\,j\in S_t}.
\ea\right.
$$
{The retailer observes the choice $j_t$.}
\item The retailer collects a random reward with expectation
\[
r_t=\begin{cases}
W_{i_tj_t} & j_t \ne 0, \\
0 & j_t = 0.
\end{cases}
\]
\end{enumerate}

\paragraph{Notation.}
For convenience, define a set of random variables $U_t$ for $t=1,2,\ldots$ independent of all variables above.
These random variables $U_t$ allow for randomized dynamic assortment personalization algorithms.
Define the filtration
\[
\mathcal F_t = \sigma(U_t,i_t,\,i_{t-1},\,j_{t-1},\,S_{t-1},\,\dots,\,i_{1},\,j_{1},\,S_{1}).
\]
$\mathcal F_t$ is the smallest $\sigma$ algebra generated by all variables known at step 2 at time $t$ 
-- it captures all information known before the retailer chooses $S_t$.
An algorithm $\pi$ is an assignment to the random variables $S_t$ so that, for every $t$,
$S_t$ is measurable with respect to $\mathcal F_t$.
Given a problem instance and an algorithm $\pi$,
let $\mathbb P^\pi$ and $\E^\pi$ be the probability and expectation measures under algorithm $\pi$:
that is, when the sets $S_t$ are selected according to algorithm $\pi$.


Define the $\Theta$-greedy algorithm, $\pi_\Theta$, which chooses
$$
S_t\in\operatornamewithlimits{argmax}_{\abs{S}\leq K}\sum_{j\in S}\frac{e^{\Theta_{i_tj}}W_{i_tj}}{1+\sum_{j'\in S}e^{\Theta_{i_tj'}}}.
$$
The $\Theta$-greedy algorithm $\pi_\Theta$ always chooses the assortment $S_t$ that would maximize revenue if $\Theta$ were the true choice parameter matrix.

\begin{definition}
Given an instance $(m,n,W,\mu^\star,\Theta^\star)$, the \emph{regret} of the algorithm $\pi$ at time $T$ is
$$
\op{Regret}(T;\pi)=\E^{\pi_{\Theta^\star}}\bracks{\sum_{t=1}^Tr_t}-\E^{\pi}\bracks{\sum_{t=1}^Tr_t}\Bigg.
$$
\end{definition}

\paragraph{Results.}
One of our main results will be to construct an algorithm $\pi_\text{nuc-norm}$ that exploits low-rank structure in $\Theta^\star$ to achieve order-lower regret compared with structure-ignorant algorithms.
\begin{theorem}[Informal]
If $\op{rank}(\Theta^\star)\leq r$ and under some additional technical conditions,
$$\op{Regret}(T;\pi_\text{nuc-norm})=O(r\max(m,n)\log T).$$
\end{theorem}
Here we see that the regret grows sublinearly in the dimension $mn$ of the problem,
for fixed rank.
We contrast this result with the best rate achievable by an algorithm that ignores the
low-rank structure of the problem.
A second result extends Theorem 1 of \citet{saure2013}
to a setting with many types.
\begin{theorem}[Informal]
If $\pi$ is a structure-ignorant algorithm and under some additional technical conditions,
$$\op{Regret}(T;\pi)=\Omega(\min(m,n)\max(m,n)\log T).$$
\end{theorem}
Turn to Theorems \ref{regcor} and \ref{regbound1} in Section~\ref{dynasssec}
for the formal statement, definition of structure-ignorant, and the technical conditions
under which the theorems hold.




\subsection{Related Work}\label{litrevsec}


Assortment personalization requires a good understanding of how consumer tastes vary.
Can a retailer learn customer preferences by observing their choices?
Discrete choice models posit answers to this question
in the form of a probability distribution over choices.
\citet{luce1959} proposed an early discrete choice model
based on an axiomatic theory, resulting in the basic attraction model.

Usually, the number of interactions
between the firm and customer is limited,
so efficient estimation of customer preferences is critical.
But estimating customer preferences is no easy task: there are combinatorially
many assortments of items, and so without further assumptions, combinatorially many quantities to estimate.
To enable tractable estimation, customer preferences are generally modeled parametrically, often using
the multinomial logit (MNL) model, which
was introduced following the work of \citet{mcfadden1974} on random utility theory.

The MNL model posits that customer choices follow a logistic model
in a vector of customer preference parameters.
Fitting a single MNL model is as simple as counting the
number of times an item is chosen relative to the other offerings.
(These counts give the maximum likelihood estimate for the model.)
The simple MNL model posits a single nominal vector of preferences
which governs the choices of all consumers.
Individual differences are modeled as random, homogeneous deviations from these universal preferences, and treated as noise.
However, these one-size-fits-all models offer no opportunity for personalization and fit heterogeneous populations poorly.

Learning a personalized model often improves performance.
A personalized model \emph{segments} the population
into \emph{types},
and fits a separate model for each type.
\bedit
In the e-commerce and brick-and-mortar settings discussed above,
the type is both \emph{discrete} and \emph{known}.
When the number of types is extremely large, this model is at least as flexible
as one which attempts to estimate a \emph{unknown} type from data.
In fact, it is possible to interpret a latent parameter matrix $\Theta$ with rank $r$
as identifying $r$ continuous features which describe each type's preferences \citep{glrm}.
Factoring the matrix of preference parameters $\Theta = UV$
with $U \in \mathbb{R}^{m \times r}$ and $V \in \mathbb{R}^{r \times n}$,
we may interpret the rows of the left factor $U$ as continuous features corresponding to each type,
from which preferences may be deduced as a linear function (by multiplying by $V$).
We distinguish these continuous features, which are implicit in our formulation,
from the explicit, discrete, and known user type.
Our model also generalizes the common latent-class model \citep{maillard2014latent}:
for example, each row of $U$ may have exactly one nonzero entry.
The index of this entry may be interpreted as indicating the customer segment \citep{glrm}.

When each type represents a single customer, the number of observations per
type (\eg, the number of distinct purchases) may be quite small.
The paucity of data on each type poses a problem for estimation methods which
require a number of observations equal to or exceeding the number of products on offer.
One solution is to aggregate customers into less fine-grained types using demographic information.
Another solution is to use methods, such as those proposed in this paper, that
require few observations per type.
One surprise in this paper is that the number of observations per type
necessary for accurate estimation may be \emph{extremely} small:
for example, simulated data shown in Figure~\ref{thefigure2} provides evidence
that a {small} number of observations per type --- a few tens --- can be sufficient,
even as the number of types $m$ and products $n$ increases!
Of course, these two approaches can also be used in conjunction:
for example, \cite{bernstein2017dynamic}
considers how to dynamically cluster customers
so as to increase the number of clusters when enough data is available.
\cite{jagabathula2017model} provides an alternative approach to combining estimation with customer segmentation.
\eedit

The MNL model has some more refined variants.
For example, the mixture of MNLs (MMNL) model models consumer choice as
a mixture of MNL models with different parameters.
With sufficiently many mixture components, an MMNL model
can approximate arbitrarily closely any choice model
that arises from a distribution over individual preferences
\citep{mcfadden2000mixed,farias2013nonparametric,van2014market}.
However, the number of interactions needed to estimate a MMNL model
is linear in the product of the number of types and the number of items.
This is an astronomical figure in most e-commerce and online advertising contexts,
where types and items both number in the millions.
{At the same time,
each user can view only so many webpages or consider only so many products ---
generally, far fewer than the number available.}
The data from each of these interactions is also limited to a single solitary choice (or lack thereof)
out of the assortment.
Our main focus in this paper is to develop a new effective method to learn and exploit
a personalized preference model despite these limits on the number of observed interactions {and the limited nature of the feedback}.

Other derivatives of the MNL model can be used to address product substitutes and complements.
These include the nested logit model \citep{williams1977} and its extensions \citep{mcfadden1980}.
Recently a new choice model was proposed that arises when substitutions from one good
to another are assumed to form a markov chain \citep{blanchet2013}.
Our paper focuses on the issue of personalization and does not consider more complex
relationships between products than is modeled by MNL; extending our methods for large-scale personalization
to these more nuanced models is a fascinating and important open challenge.


Choosing the optimal assortment can be computationally hard or computationally
easy depending on the choice model.
Under the MNL model, it is easy to optimize assortments:
\citet{talluri2006} show that presenting items in revenue sorted order is always optimal.
On the other hand, it is NP-hard to optimize a single assortment
to be offered to one MMNL population, even with only two mixture components,
but approximation schemes exist
\citep{rusmevichientong2014}.
Assortment optimization over the nested MNL model is computationally hard
in general \citep{davis2014} but easy in some cases \citep{li2015d}.
Optimizing an assortment of constrained cardinality under the MNL model
is easy \citep{megiddo1979combinatorial,rusmevichientong2010},
while optimizing an assortment with weighted budget constraint is hard
\citep{desir2014near}.

Assortment optimization with limited inventory is even more complex. Such problems
need to be solved over multiple periods, a different assortment offered each period
to take into account possible future stockouts. \citet{talluri2006} solve the classic
problem with a single MNL model, while \citet{bernstein2011dynamic,golrezaei2014real}
solve a corresponding problem with a mixture model, representing a few customer segments. In all of
these, all preferences of all populations are assumed known.


When preferences are unknown and are to be learned simultaneously with assortment optimization,
we can conceive of assortments as bandit arms and consumer choice as bandit feedback to get the problem
of \emph{dynamic assortment optimization} (without context).
\citet{rusmevichientong2010} formulated
this problem
when choices are governed by a single MNL model and showed that their algorithm has regret upper bounded in order
by $n\log^2 T$ for $n$ items and $T$ interactions. \citet{saure2013} improved this to $n\log T$ and showed that
this order is optimal. Earlier, \citet{caro2007dynamic} were the first to conceive of dynamic optimization of assortments under learning by studying a related but different problem. Their problem differs in that the demand to be learned is assumed exogenous and independent of the combination offered and other items.
The assortments constitute a simultaneous play of multiple arms, rather than an optimal assortment from which a customer chooses zero or one items.

Our \emph{dynamic assortment personalization} is an instance of a \emph{contextual} bandit problem with discrete contexts. It arises when at each interaction, a different context, drawn from some finite set, is observed. Based on this discrete contextual information,
the problem is to personalize the assortment to target each context as well as possible, while also learning to improve performance.
In particular, a good algorithm will use lessons learned in
one context to improve performance in other contexts.
To the best of our knowledge, we are the first to consider this dynamic assortment personalization problem,
and in particular the first to consider any stochastic bandit with {discrete} contexts, rather than continuous contexts with a functional relationship to rewards (such as linear).
\citet{lai1985asymptotically} posed the classic stochastic multi-armed bandit problem, in which each of $n$ arms
has an initially-unknown bounded reward distribution and in each
time step one chooses one arm to pull
with the overall goal of minimal regret: the expected difference in reward between the prescient policy that always pulls the best arm and one's actual performance. An alternative formulation of the multi-armed bandit problem
involves rewards that, instead of being distributed according to a fixed unknown distribution, may change adversarially in response to the choice of arm. For a discussion of the differences between stochastic and adversarial bandits we refer the reader to \citet{bubeck2012regret}. In this paper we focus solely on the \emph{stochastic} bandit. Examples of contextual stochastic bandits include
\citet{rigollet2010nonparametric,perchet2013multi,goldenshluger2013linear,slivkins2014contextual,bastani2015online}, which all focus on the setting with $n$ generally unrelated arms, where each arm is associated with a regression function that governs the expected reward conditioned on a continuous vector of covariates representing context. The former two papers assume a general non-parametric functional dependence; the latter three assume a linear regression function. In all these papers, the context
is parametrized by a continuous (scalar or vector) quantity;
in other words, the relation between different contexts is embedded topologically and known in advance.

In contrast, in the dynamic assortment personalization, the relation between different contexts must be learned from the data.
Contexts are discrete; they correspond to rows of an unknown parameter matrix which governs consumer choice.
The observation of choice can be likened (imperfectly) to the noisy observation
of an entry of the matrix. This analogy brings to mind the problem
of matrix completion: the problem of (approximately) recovering
an (approximately) low rank matrix
from a few (noisy) samples from its values.

\beditb
\cite{glrm} consider how to optimize the convex losses that arise in a general class of entry-wise matrix observation models, such as noisy observation of a few entries of a low-rank matrix.
The conditional MNL choice model developed in the present paper moves beyond the models considered in \cite{glrm},
as each observation depends on \emph{several} entries in the parameter matrix.
Furthermore, we develop new statistical guarantees and dynamic extensions that
are beyond the scope of \citep{glrm}.
\eeditb

Some of our results are in the same vein as statistical matrix
completion bounds.
Following groundbreaking work on exact completion
of exactly low rank matrices
whose entries are observed without noise
\citep{candes2010,candes2009,recht2010,keshavan2010},
approximate recovery results have been obtained for
a variety of different noisy observation models.
These include observations with additive gaussian \citep{plan2009}
and subgaussian \citep{keshavan2009} noise,
0-1 (Bernoulli) observations \citep{davenport2014},
observations from any exponential family distribution \citep{gunasekar2014},
and observations generated according to
the Bradley-Terry-Luce model for pairwise comparisons \citep{lu2014,oh2015}.
\bedit
These are most related to our Theorem~\ref{mainthm}, which explores
the \emph{static} estimation problem.

Our Theorem~\ref{mainthm}
differs from previous work on matrix recovery in at
least three critical ways.
\beditb
First, the data for our problem consists of choices from an assortment,
rather than entrywise observations of ratings, pairwise comparisons, or full rankings;
second, Theorem~\ref{mainthm} holds when assortments are subsets of
the full set of items,
rather than chosen iid with replacement so that duplicate items sometimes appear;
and third, customers in our model can choose not to purchase any item from the presented assortment.
In sum, Theorem~\ref{mainthm} shows how high dimensional parameters can be recovered from a sublinear number of transactional observations of consumer choice.
\eeditb

\beditb
A variety of techniques have been developed
in the literature to analyze
the recovery of high dimensional parameters using regularized maximum likelihood.
Our proof of Theorem~\ref{mainthm} uses the machinery of restricted
strong convexity originally developed by \cite{negahban2011,negahban2012unified} and applied to the case of noisy matrix completion and other entrywise observation models.
The present paper leverages this machinery and extends it to our setting, where we observe assortment choices
rather than individual entries of the matrix of preference parameters,
and thereby moves beyond the sorts of entrywise models analyzed in \cite{negahban2011,negahban2012unified}.
As reviewed in Section~\ref{proofsketch}, because observations depend on \emph{several} entries of the parameter matrix,
this extension relies on a \emph{multivariate} Rademacher
comparison lemma \cite[Lemma 7]{bertsimas2014predictive} to establish the desired restricted strong convexity.
This lemma occurred as a technical lemma in an unrelated result in
 \cite{bertsimas2014predictive}, which studies
 conditional stochastic optimization given continuous contextual information using local reweighting schemes.
Beyond Theorem~\ref{mainthm}, which studies the recovery of $\Theta^*$
from a static dataset, this paper further shows how to use this static result in a dynamical setting.
In Section~\ref{mainprobsec}, we describe the
high-dimensional contextual dynamic assortment personalization problem
for which this paper provides a novel algorithm and analysis.
\eeditb

\eedit

\section{The Low-Rank {Conditionally} Multinomial Logit Choice Model}\label{s-model}

In this section we describe the {low-rank conditionally multinomial logit choice (LRCMNL) model,} study the static estimation problem under
observing only choice, propose an estimator, prove recovery bounds, and develop a fast
algorithm for computing the estimator from large-scale data.

The {conditionally} MNL (CMNL) model over types $i=1,\dots,m$ and items $j=1,\dots,n$
is parameterized by $\mu^\star\in\Delta^m$ and $\Theta^\star\in\R{m\times n}$
and describes two random variables: type $I$ and choice $J$.
Type $I$ is assumed to be distribution according to $$\Prb{I=i}  = \mu^\star_i.$$
For any given assortment $S\subset\braces{1,\dots,n}$, choice $J$
is assumed to be distributed according to the following model
\begin{align}\label{themodel}\begin{split}
\Prb{J = j ; S} &= \sum_{i=1}^n\Prb{I=i}\Prb{J = j \mid I = i; S}\\
\Prb{J = j \mid I = i; S} &=
\frac{1}{\sum_{j' \in S} e^{-\Theta^\star_{ij'}}}\times
\left\{\ba{lc}
1&\quad\quad{j=0}\\
0&\quad\quad{j\neq0,\,j\notin S_t}\\
\exp(\Theta^\star_{i_tj})&\quad\quad{j\neq0,\,j\in S_t}
\ea\right.
\end{split}\end{align}
where $J=0$ represents the choice not to choose from $S$ -- an option that is always available for any assortment $S$.

The {LRCMNL} model posits a {CMNL} model in which the parameter $\Theta^\star$ has low rank:
$$\op{rank}(\Theta^\star)\ll m,n.$$
We will also consider the case where $\Theta^\star$ has approximately low rank:
$$r\ll m,n,\quad\overline{\sigma}_{r+1}(\Theta^\star)\approx 0,$$
where $\overline{\sigma}_{r+1}(\Theta^\star)$ is the sum of the singular values of $\Theta^\star$ smaller than the $r\thh$ largest singular value, i.e.,
\begin{equation}\label{remainingsingularvalues}
\overline{\sigma}_{r+1}(\Theta^\star)=\sum_{j=r+1}^{\min\{m,n\}}\sigma_{j}(\Theta^\star).
\end{equation}

\subsection{Implications of the {LRCMNL} Model}

Let us consider when the choice distribution should follow a {LRCMNL} model.

Suppose that each individual in the population makes
\emph{rational} choices:
that is, choices maximize utility with respect to a vector of
utilities randomly distributed over the population.
This is called a random utility choice model.
The approximation results of \citet{mcfadden2000mixed} show that
for any random utility choice model,
there is a variable $I$ such that the choice distribution
is approximately MNL conditioned on $I$.
{If $I$ is an observed variable, then this corresponds to a CMNL model.
More generally, this result suggests that for any sufficiently fine partition of types $I$,
the CMNL model forms a reasonable approximation of the choice model.
That is, while a large population may have a complex choice model due to heterogeneity of individuals,
the MNL choice model should provide a good approximation for the decisions made by a single individual.}

Conversely, any {CMNL} model, including the {LCMMNL} model,
{inherits an interpretation as a random utility choice model from the MNL model.}
Let 
${\Theta^\star_{ij}}$
be the mean utility type $i$ enjoys from item $j$.
Let us suppose that the utility of each customer of type $i$ is the sum of the
mean utility of type $i$ together with a random idiosyncrasy distributed
according to the Gumbell (extreme value) distribution, and that each customer
chooses an item by maximizing her utility among the items on offer:
\begin{equation}\label{RUM}
J=\max_{j\in S_t}\prns{\Theta^\star_{Ij}+\zeta_{j}}\quad\text{where}\quad\zeta_{j}\sim\op{Gumbell}(0,1).
\end{equation}

The {LRCMNL} model \eqref{themodel} can therefore arise in either of two ways.
It describes choice behavior when customers are clustered into types within each of which
customers have a private, idiosyncratic utility distributed as in \eqref{RUM}
and the heterogeneity of the population is described by the varying
mean utilities {$\Theta_{ij}$} over types $i$.
The {LRCMNL} model also describes choice behavior when each customer is her own type.
The random idiosyncrasies associated with each choice event
reflect human inconsistencies in decision making
or slight variations over time in preferences
\citep{kahneman1979prospect,deshazo2002designing}.

Learning the preferences of multiple, heterogeneous customer types
simultaneously is difficult without additional structure.
Both \citet{bernstein2011dynamic} and \citet{golrezaei2014real} study multi-period assortment
optimization problems with multiple, heterogeneous customer types
assuming full knowledge of the distribution of consumer choice.
Both undertake case studies in which they estimate these distributions from static data
in order to evaluate the performance of their optimization
algorithms on distributions
that mimic real data. However, in both cases, they allow only a few
segments (3 and 10, respectively). One reason for this choice may be that
estimation becomes intractable for models with many more segments.
Our model, by contrast, can tractably estimate distributions with large numbers of types and items.
We overcome limitations of previous models by assuming that
the underlying dimension of the model is small
in the sense that our parameter matrix has (approximate) low rank.

If $\Theta^\star$ has (approximate) rank $r$, we may factor $\Theta^\star$ to find
vectors $u_{i\ell},\,v_{j\ell}\in\Rl$ for $i=1,\dots,m,\,j=1,\dots,n,\,\ell=1,\dots,r$
such that $\Theta^\star_{ij}$ is (approximately)
equal to $\sum_{\ell=1}^ru_{i\ell}v_{j\ell}$. The right factors $v_{j\ell}$
can be thought of as latent item features, and the left factors $u_{i\ell}$ as
latent type weights which characterize how much type $i$ values feature $\ell$.
When $\Theta^\star$ has (approximate) low rank, we can be sure that just
a few latent features suffice to (approximately) explain consumer choice,
and these latent features need not be measurable or have a physical
interpretation.
Indeed, the number of features that matter for decision making may be constrained by cognitive load:
to consider many features would require proportional time and energy.
However, even if consumer utility is a non-linear function of item features,
and even if the number of features required to describe an item is extremely large,
\cite{udell2017nice} prove that large enough preference matrices are still approximately low rank
so long as types and items are drawn iid from some population.

In summary, the {LRCMNL} model is implied by the assumption that choice is rational with a utility distribution with means that depend on only a few (possibly unknown) features.
Usually very few features suffice
due to the finite range of human perception and rationality
or simply due to concentration of measure.

\subsection{The Static Estimation Problem}\label{obsmodsec}

Next, we describe an observation model and the problem of estimating the {LRCMNL} parameters from observed data.
We suppose that we have $N$
observations $\{(i_t,\,j_t,\,S_t):t=1,\dots,N\}$ where $S_t$ is sampled uniformly at random from the set of subsets of $\{1,\dots,n\}$ of size $K_t$, the sequence $K_t$ is arbitrary (possibly random) satisfying $K_t \leq K$, and $i_t,j_t$ are iid according to the model \eqref{themodel}.
We also assume that $\magd{\Theta^\star}_\infty\leq\alpha/\sqrt{mn}$ for purely technical reasons.
The assumption of (normalized) bounded entries assumption is standard in many matrix completion recovery results (see Section \ref{litrevsec}) and is necessary for our proof of recovery with high probability; see below.

It is important to highlight that our observation model consists of observing only the choice made by customers.
In practical applications, this is typically the only observation possible.
Moreover, it is generally truthful since it is utility maximizing, unlike reporting rankings in a survey or focus group.

\subsubsection{Our Estimator}
Define the negative log likelihood of the observations
given parameter $\Theta$ as
\begin{equation}\label{negloglikeq}
L(\Theta)=\frac1N\sum_{t=1}^N\log\prns{\prns{1+\sum_{j\in S_t}e^{\Theta_{ij}}}\prns{\left\{\ba{lc}1 \quad\quad& j_t=0\\e^{\Theta_{ij_t}} \quad\quad& \text{otherwise}\ea\right.}^{-1}}.
\end{equation}
We define our estimator $\widehat \Theta$ for $\Theta^\star$ as any solution of
the nuclear norm regularized maximum likelihood problem
\begin{equation}
\label{eq-alg}
\ba{rl}
\mbox{minimize} & L(\Theta) + \lambda\|\Theta\|_*, \\
\mbox{subject to} & \|\Theta\|_\infty\leq\alpha/\sqrt{mn},
\ea
\end{equation}
where $\lambda > 0 $ is a tuning parameter and the nuclear norm $\|\Theta\|_*$
is the sum of the singular values of $\Theta$.
We use $\widehat{\Theta}$ to denote the solution to this problem.

The constraint $\|\Theta\|_\infty\leq\alpha/\sqrt{mn}$
appears purely as an artifact of the proof; we recommend to omit
this constraint in practice. We omit this constraint both in our
specialized algorithm (Section \ref{fgdalgorithm}) and
in our numerical results (Section \ref{numerics});
the good practical performance on examples demonstrates
the practical irrelevance of this constraint.

Problem~\eqref{eq-alg} is convex and hence can be solved by a variety of standard convex methods that take advantage of the special structure of the problem \citep{cai2010singular,parikh2014proximal,hazan2008sparse,orabona2012prisma}.
In Section \ref{fgdalgorithm} we provide a specialized first-order algorithm that, in fact,
works on the \textit{non-convex}, factored form of the problem for increased speed,
but still guarantees convergence to the global optimum
with high probability. 

Our estimator $\hat\mu$ for the customer type distribution $\mu^\star$ is the empirical frequencies of each type:
$$\hat\mu_i=\frac1N\sum_{t=1}^N\indic{i_t=i}.$$

\subsubsection{Recovery Guarantee for the Parameter Matrix}

In this section, we bound the error of the estimator $\widehat\Theta$.
Our bound depends on the following quantities,
which capture the complexity of learning the preferences of all customer types over all items.
\begin{itemize}
\item \emph{Number of observations.} The bound decreases as the number $N$ of observations increases.
\item \emph{Number of parameters.} The bound grows with the dimensions $m,\,n$ of the parameter matrix $\Theta^\star$.
\item \emph{Underlying rank dimension.} For any $r \leq \min\prns{m,n}$, our bound decomposes into two error terms.
The first error term is the error in estimating the top $r$ ``principal components'' of the parameter matrix.
This error term grows with $\sqrt{r}$ and captures the benefit of learning only the most salient features instead of all parameters at once.
The second error term is the error in approximating the parameter matrix by only its top $r$ ``principal components.''
In particular, if $\Theta^\star$ is exactly rank $r$, then this second error term vanishes.
More generally, however, we may be interested in estimating parameter matrices that are only approximately low rank,
\ie, with quickly decaying singular values past the top $r$.
In this case, our bound depends on the sum of the remaining singular values.
\item \emph{Size of parameters.} Our bound grows with the (scaled) maximum magnitude of any entry
$\alpha$.
\item \emph{Size of assortments.} Our bound grows with the maximum
size $K$ of the assortments.
\end{itemize}

\begin{theorem}\label{mainthm}\beditb
Let $\tau\geq1$ be given,
$\rho\geq1$ be such that $1/\rho\leq m\mu_i\leq \rho$ $\forall i=1,\dots,m$, and
$\alpha$ be such that $\magd{\Theta^\star}_\infty\leq\alpha/\sqrt{mn}$.
Fix $\lambda = 8\sqrt{\frac{\tau\rho K(m+n)\log (m+n)}{mnN}}$.
Suppose $N\leq mn\log (m+n)$.
Then under the observation model in Section \ref{obsmodsec} and
for any integer $r\leq \min\braces{m,n}$,
with probability at least $1-3(m+n)^{-\tau}$,
any solution $\widehat \Theta$ to Problem~(\ref{eq-alg}) satisfies
\begin{align*}
\|{\widehat\Theta-\Theta^\star}\|_\text{F}\leq
4096\sqrt{\tau}\alpha e^{\frac{8\alpha}{\sqrt{mn}}}
\max\Biggl\{&
\sqrt{\frac{r(m+n)\log(m+n)}{N}(\rho K)^3},\,\\
&\quad\quad\quad\quad\prns{\frac{\overline{\sigma}_{r+1}(\Theta^\star)(m+n)\log(m+n)}{N}(\rho K)^3}^{1/4}
\Biggr\},
\end{align*}
where $\overline{\sigma}_{r+1}(\Theta^\star)$ is the sum of the remaining singular values of $\Theta^\star$ smaller than the $r\thh$ largest singular value, as defined in eq. \eqref{remainingsingularvalues}.\eeditb
\end{theorem}

A few remarks on this theorem are in order.
\begin{itemize}
  \item If $\Theta^\star$ were exactly low rank ($\overline{\sigma}_{r+1}(\Theta^\star)=0$), then a number of observations scaling slightly faster than $r\max(m,n)\log(m+n)$ are needed in order to obtain a consistent estimate for
  $\Theta^\star$. {That is to say, if the number of products
  is growing no faster than the number of types $n=O(m)$ and rank is bounded $r=O(1)$, then the \emph{number of observations per type},
  $N/m$,
  needed in order to estimate everyone's preferences consistently is
  logarithmic, $N/m=\omega(\log(m))$.
  }
  \item The first term in the bound represents the \emph{estimation error}: the difficulty of estimating the top rank-$r$ approximation to $\Theta$ from only $N$ samples.
  The second term in the bound represents the \emph{approximation error} in the model:
  the error incurred because the target rank $r$ is smaller than the true rank of $\Theta^\star$. This term is zero when $\op{rank}(\Theta^\star)\leq r$ and is small when the singular values of $\Theta^\star$ that are smaller than the $r\thh$ smallest one are small.
  \item The choice of $\lambda$ does not depend on $r$ and the result holds for any $r\leq\min\braces{m,n}$. That means that it is not necessary to know the rank or approximate rank of $\Theta^\star$ -- as long as it has (approximate) low rank for some unknown but not too large $r$, our algorithm will be able to recover $\Theta^\star$ with high fidelity.
  \item The proof of this theorem requires a bound $N\leq mn\log (m+n)$ on the maximum number of
  observations used to fit the estimator. From a practical perspective, this upper bound presents no difficulties: generally, the estimation problem is hard when few observations are available; whereas when $N > mn\log (m+n)$ simpler approaches such as maximum likelihood estimation can perform well and give consistent estimates.
  Furthermore, in high-dimensional settings it is generally impossible to
  exhaustively sample all $mn$ type-item pairs; hence as a practical matter,
  we will always have that $N\leq mn\log(m+n)$ holds. {We discuss
  this at greater length in the dynamic setting in Section~\ref{s-two-phase}.}
  \item The parameter $\tau$ controls the probability of the result.
  Choosing $\tau=1$, we see the theorem already holds with extremely high probability, which converges to $1$ as either $m$ or $n$ grow.
  Other values for $\tau$ give greater generality to the theorem.
  We will see a more sophisticated use of this probability control $\tau$ in the proof of Theorem~\ref{regbound1}.
\end{itemize}

A closely related result to Theorem \ref{mainthm} appeared in our preliminary work \citep{kallus2015learning}, which focuses only on the static estimation problem and only in the absence of the no-choice option.
In assortment personalization, we must consider estimation in the permanent presence of a no-choice option in \emph{any} assortment and where the mixtures $\mu$ are not necessarily uniform, and we must also consider the decision problem involved in dynamically offering personalized assortments. None of these appear in our brief preliminary work.
\beditb
\subsubsection{Proof sketch for Theorem~\ref{mainthm}}\label{proofsketch}
In low dimensions,
a standard proof that the minimizer of a loss function
is close to the true parameters shows
1) the loss function is strongly convex, and
2) the the true parameters achieve low loss.
Since the loss is strongly convex, any near-optimal parameter must be close in Euclidean distance to the optimal one.
However, the loss function we use is \emph{not} strongly convex,
nor can it be until every item has been offered to every type.

Instead, we argue as follows.
1) The nuclear norm of the error $\Delta = \widehat \Theta - \Theta^\star$
controls its square Frobenius norm.
The proof of this uses random sampling to show that
the Bregman divergence of the loss function $L(\Theta)$ is (with high probability)
strongly convex around $\Theta^\star$,
and uses the form of the objective to bound this Bregman divergence above by the nuclear norm.
2) The Frobenius norm of the low rank matrix $\Delta$ controls its nuclear norm.
We combine these statements to bound the Frobenius norm of the error $\Delta$.

The three key steps in our proof use the three important ingredients in our method:
random sampling, regularized empirical risk minimization, and an approximately low rank parameter matrix.
\begin{enumerate}[start=0,leftmargin=*,labelindent=0in,align=left]
  \item Our proof fundamentally relies on establishing the following inequality
  \begin{equation} \label{eq-fro-lt-nuc-explain}
  \magd{\Delta}_\text{F}^2\leq 256\sqrt{\tau}\alpha e^{\frac{8\alpha}{\sqrt{mn}}} (\rho K)^{3/2}\sqrt{\frac{(m+n)\log (m+n)}{N}}\magd{\Delta}_*,
  \end{equation}
  which shows that the nuclear norm of the error $\Delta$ controls the square Frobenius norm of the error.
  We begin our proof by defining a set $\mathcal A^\star$ so that
  eq.~\eqref{eq-fro-lt-nuc-explain} holds by definition for any $\Delta \not \in \mathcal A^\star$:
  $$\mathcal A^\star=\braces{\Delta:~
\|\Delta\|_\infty\leq\frac{2\alpha}{\sqrt{mn}},\,
\|\Delta\|_F^2\geq{{\max\braces{(18\tau)^{1/4},\,480}}}{{\rho^{3/2}K^{1/2}}\alpha}\sqrt{\frac{\sqrt{mn}\log (m+n)}{N}}\|\Delta\|_*}.$$
(Recall $\|\Delta\|_\infty\leq\frac{2\alpha}{\sqrt{mn}}$ holds by assumption.)

The rest of our proof will show that eq.~\eqref{eq-fro-lt-nuc-explain} holds,
with high probability, even when $\Delta \in \mathcal A^\star$.
  \item
  Our main task is to show restricted strong convexity:
  our loss function is strongly convex
  around $\Theta^\star$ when $\Delta$ is restricted to $\mathcal A^\star$.
  This is established by proving that
  (with high probability, for every $\Delta \in \mathcal A^\star$)
  the Bregman divergence of our loss function,
  \[
  D_{\Theta^\star}(\Delta) = L(\Theta^\star + \Delta)-L(\Theta^\star)-\nabla L(\Theta^\star)\cdot \Delta,
  \]
  bounds (a constant times) the square Frobenius error $\|\Delta\|_\text{F}^2$.
\begin{enumerate}[leftmargin=*,labelindent=0in,align=left]
  \item We first prove that this bound holds in expectation.
  In Lemma~\ref{lem:quad-lower}, we use Taylor's theorem to show
  $$
  D_{\Theta^\star}(\Delta)\geq\frac1{4Ke^{8\alpha/\sqrt{mn}}}\frac1N\sum_{t=1}^NY_t(\Delta),\quad\text{where}\quad Y_t(\Delta)=\frac1{K_t}\sum_{j_t\in S_t}\Delta_{i_tj_t}^2.
  $$
  The expectation of the empirical process $\frac1N\sum_{t=1}^NY_t(\Delta)$
  is at least $\frac1{\rho m n}\|\Delta\|_\text{F}^2$ because
  the observed sets $S_t$ are \textbf{sampled randomly}.

  \item The second step, and the key to our proof, is to show that this process concentrates
    \emph{uniformly} around its expectation for all $\Delta\in\mathcal A^\star$.
    A major difficulty arises due to our particular observation model:
    $Y_t(\Delta)$ depends on several entries of $\Delta$ at once.
    This dependence is key to modeling a realistic e-commerce setting
    where customers do not purchase more than one similar item and
    where customers retain the option not to purchase any item at all.
    We are not aware of other related work that can handle this dependence.
    For example, \citep{oh2015} avoid the problem by sampling items
    {with replacement} and {without a no-purchase option}.

    Our proof proceeds as follows. We first peel $\mathcal A^\star$ by intersecting
    it with concentric spherical shells: 
  $\mathcal A^\star=\{0\}\cup{\bigcup_{l=1}^\infty\mathcal A_l}$ where
  $\mathcal A_l=\braces{\Delta\in\mathcal A:\eta\beta^{l-1}\leq\magd{\Delta}_\text{F}\leq\eta\beta^{l}}$, $\eta=\inf_{\Delta\in\mathcal A^\star:\Delta\neq0}\magd{\Delta}_\text{F}>0$, and $\beta>1$.
  In Lemma~\ref{lem:strongly-convex-bounded}, we study in each peel separately the maximal deviations of the empirical process from its expectation,
  $$\mathcal M_l=\sup_{\Delta\in\mathcal A_l}\prns{\frac1N\sum_{t=1}^NY_t(\Delta)-\E\frac1N\sum_{t=1}^NY_t(\Delta)}.$$
  We first show that $\mathcal M_l$ is itself concentrated near its expectation, so we need only bound its expectation.
  Using a symmetrization argument, we let $\epsilon_t$ be iid Rademacher variables (equiprobably $\pm1$) and show that
  $\E\mathcal M_l\leq2\E\sup_{\Delta\in\mathcal A_l}\frac1N\sum_{t=1}^N\epsilon_t{Y_t(\Delta)}$,
  known as the Rademacher complexity.
  Computing this quantity is made difficult because a) $Y_t(\Delta)$ depends on multiple elements of $\Delta$ and b) the identity of these elements is not independent because sampling is without replacement. We therefore use a multivariate Rademacher comparison lemma \cite[Lemma 7]{bertsimas2014predictive} in order to prove that
  $$\E\mathcal M_l\leq2\E\sup_{\Delta\in\mathcal A_l}\frac1N\sum_{t=1}^N\epsilon_t{Y_t(\Delta)}\leq 4\magd{\Delta}_\infty\E\sup_{\Delta\in\mathcal A_l}\frac1N\sum_{t=1}^N\sum_{j\in S_t}\epsilon_{tj}{\Delta_{ij}},$$
  where $\epsilon_{tj}$ are new iid Rademacher variables. This latter complexity is much simpler and amenable to analysis. Using the concentration of random matrices \citep{tropp2012user}, we control it by the norm of $\Delta$, which is bounded by the peeling construction.


  \item In Lemma~\ref{lem:strongly-convex}, we use a union bound over the peels
  to obtain the desired restricted strong convexity:
  with high probability,
  \begin{equation}\label{strong-convex-explain}
  D_{\Theta^\star}(\Delta)\geq
  \frac1{8Ke^{8\alpha/\sqrt{mn}}\rho mn}
  \magd{\Delta}_\text{F}^2,\quad\forall \Delta\in\mathcal A^\star.
  \end{equation}
\end{enumerate}
  \item In Lemma~\ref{regularizedobjectivelemma}, we show that our choice of \textbf{nuclear-norm-regularized log likelihood objective} allows us to bound the Bregman divergence above as follows
  $$D_{\Theta^\star}(\Delta)\leq\prns{\magd{\nabla L(\Theta^\star)}_2+\lambda}\magd{\Delta}_*.$$
  This quantity is controlled by the optimality of $\Theta^\star$ in the log likelihood objective in terms of its first-order condition.
  In Lemma~\ref{lem:truth-near-opt}, by leveraging random matrix concentration bounds \citep{tropp2012user},
  we show that $\Theta^\star$ is near optimal with high probability:
  \[
  \magd{\nabla L(\Theta^\star)}_2\leq 4\sqrt{\tau}\sqrt{\frac{\rho K(m+n)\log (m+n)}{mnN}}\leq\lambda/2,
  \]
  where the last inequality is by our choice of $\lambda$.
  Combining $D_{\Theta^\star}(\Delta)\leq2\lambda$, our choice of $\lambda$, and eq.~\eqref{strong-convex-explain} shows that eq.~\eqref{eq-fro-lt-nuc-explain} holds even for $\Delta\in\mathcal A^\star$ (with high probability).

  \item Finally, in Lemma~\ref{spectraldecompositionlemma}, using a spectral decomposition argument, we show that, if $\Theta^\star$ is \textbf{low rank or approximately low rank} and also near-optimal in that $\magd{\nabla L(\Theta^\star)}_2\leq\lambda/2$, then
  the nuclear norm of $\Delta$ is tightly controlled by the Frobenius norm of $\Delta$:
  \begin{equation} \label{eq-max-frob-or-rho-explain}
  \|\Delta\|_* \leq 16\max\braces{\sqrt{r}\magd{\Delta}_\text{F},\|{\overline\Theta^\star_r}\|_*},
  \end{equation}

\end{enumerate}
Combining eqs.~\eqref{eq-fro-lt-nuc-explain} and \eqref{eq-max-frob-or-rho-explain} yields our main result.
\eeditb

\subsubsection{Recovery Guarantee for the Customer Type Distribution}

In this section, we bound the error in our estimator $\hat\mu$
for the true customer type distribution $\mu^\star$.

\begin{theorem}\label{muthm}
Let $\tau\geq0$ and $q\in[1,\infty]$ be given. With probability at least $1-e^{-\tau}$,
$$
\magd{\hat\mu-\mu^\star}_q\leq \min\braces{8\sqrt{\frac{\tau+m}{N}},~\frac{m^{1/q}}{\sqrt2}\sqrt{\frac{\tau+\log(2m)}{N}}}.
$$
\end{theorem}

The second term in the above min is immediate from applying Hoeffding's inequality to each component and using the union bound. For $q=1$ and any $\tau$, however, the Hoeffding-based bound diverges for any $N=O(m^2)$ (in general it goes to zero only for $N=\omega(m^{2/q})$). We derive the first term using a Rademacher complexity argument. The first term goes to zero for any $q$ as long as $N=\omega(m)$ grows superlinearly in $m$. This first term is critical for showing that $\mu^\star$ can be estimated consistently in the $q$ norm for $q<2$. In particular, the theorem provides for $\sqrt{m/N}$-consistent estimation of the type distribution in the $\ell_1$ norm.


\subsection{A Factored Gradient Descent Algorithm}\label{fgdalgorithm}

In this section, we develop a specialized first-order algorithm for computing $\widehat\Theta$ that works
on the non-convex, factored form of the nuclear-norm regularized likelihood optimization problem. The algorithm
is particularly economical with memory because (a) it does not keep all $m\times n$ optimization variables in memory but rather only $\tilde r\times (m+n)$ where $\tilde r$ is a guess at rank and (b) it eschews the use of \emph{any} spectral computation such as SVD (or partial SVD) at each iteration. This makes the algorithm particularly useful for
the large scale data encountered in e-commerce applications.

Our factored gradient descent (FGD) algorithm
solves the problem
\begin{equation}
\label{eq-alg-alt}
\ba{rl}
\mbox{minimize} & L(\Theta) + \lambda\|\Theta\|_*.\ea
\end{equation}
As discussed in Section~\ref{obsmodsec},
the algorithm we employ does \emph{not} enforce any constraint on $\magd{\Theta}_\infty$.
A constraint of this form is necessary for the technical result in our main
theorem, but is unnecessary in practice, as can be seen in our numerical results in Section~\ref{numerics}.

In applications, one is interested in solving the problem \eqref{eq-alg-alt}
for very large $m,\,n,\,N$. Due to the complexity of Cholesky factorization,
this rules out theoretically-tractable second-order interior point methods.
One standard approach is to use a first-order method, such as
\citet{cai2010singular,parikh2014proximal,hazan2008sparse,orabona2012prisma};
however, this approach requires (at least a partial) SVD at each step.
An alternative approach, which we take here,
is to optimize as variables the factors $U\in\R{m\times \tilde r}$ and $V\in\R{n\times \tilde r}$
of the optimization variable $\Theta=UV^T$ rather than producing these
via SVD at each step; see, \eg, \citet{keshavan2009matrix, jain2013low}.
To guarantee equivalence of the problems, we must take $\tilde r=\min\prns{m,n}$.
However, if we believe the solution is low rank
or if we want to enforce low rank, then
we may use a smaller $\tilde r$, reducing computational work and storage.

Our FGD algorithm proceeds by applying gradient descent steps to the unconstrained problem
\begin{equation}
\label{eq-alg-alt2}
\ba{rl}
\mbox{minimize} & L(UV^T) + \frac\lambda2\|U\|_\text{F}^2 + \frac\lambda2\|V\|_\text{F}^2,\\
\mbox{subject to} & U\in\R{m\times \tilde r},\,V\in\R{n\times \tilde r}.\ea
\end{equation}
This formulation has the advantage of being unconstrained, with a differentiable
objective function, making it amenable to solution via simple optimization methods
\citep{recht2010, recht2011hogwild, glrm}.
\begin{lemma}\label{equivprobs}
Problem \eqref{eq-alg-alt2} is equivalent to
\begin{equation}
\label{eq-alg-alt-rank}
\ba{rl}
\mbox{\emph{minimize}} & L(\Theta) + \lambda\|\Theta\|_*\\
\mbox{\emph{subject to}} & \op{rank}(\Theta) \leq \tilde r.
\ea
\end{equation}
\end{lemma}
That is, Problem \eqref{eq-alg-alt2} is equivalent to
Problem \eqref{eq-alg-alt} subject to an additional rank constraint $\op{rank}(\Theta)\leq\tilde r$.
If, for a particular choice of loss function $L$, Problem \eqref{eq-alg-alt}
has a solution with rank less than $\tilde r$, then
the rank constraint is not binding, so Problem \eqref{eq-alg-alt} is itself
equivalent to Problem \eqref{eq-alg-alt2}.
(Recall that we defer all proofs to the Appendix.)

It is easy to compute the the gradients of the objective of \eqref{eq-alg-alt2}.
Since $L(\Theta)$ is differentiable,
\begin{align*}
\nabla_U L(UV^T) &= \nabla L(UV^T)V,\\
\nabla_V L(UV^T) &= \nabla L(UV^T)^TU.
\end{align*}
We do not need to explicitly form $\nabla L(UV^T)$ in order to compute
these;
computing gradients implicitly reduces the memory required to implement the algorithm
(see Algorithm \ref{alg:fgd}).
Similarly, we need not form $UV^T$ to compute $L(UV^T)$.
Recent work
has shown that gradient descent on the factors 
converges linearly to the global optimum
for problems that enjoy restricted strong convexity \citep{bhojanapalli2015dropping}.
In eq. \eqref{eq-rsc} in the proof of Theorem \ref{mainthm}
we establish restricted strong convexity for our problem with high probability.
Hence with high probability, FGD converges to the global minimum of
Problem \eqref{eq-alg-alt2}, and hence to the global minimum of
Problem \eqref{eq-alg-alt} provided $\tilde r$ is chosen to be large enough.

\begin{algorithm}[t!]\SingleSpacedXI
   \caption{Factored Gradient Descent for \eqref{eq-alg-alt}}
   \label{alg:fgd}
\begin{algorithmic}
   \State {\bfseries input:} {dimensions $m,\,n,\,\tilde r$, data $\{(i_t,j_t,S_t)\}_{t=1}^N$, regularizing coefficient $\lambda$, and tolerance $\tau$}
   \State $U \leftarrow U^0$,
          $V \leftarrow V^0$, $f' \leftarrow \infty$.
   \Repeat
   \State $\eta \leftarrow 1,\,f\leftarrow f',\,\Delta U\leftarrow -\lambda U,\,\Delta V\leftarrow -\lambda V$
   \For{$t=1,\dots,N$}
       \State $Q=0$
       \For{$j\in S_t$}
           \State $w_j\leftarrow e^{- U_{i_t}^T V_j}$
           \State $W\leftarrow W+w_j$
       \EndFor
       \State $\Delta U\leftarrow \Delta U - \frac{1}{N} \prns{e_{i_t}V_{j_t}^T-\frac{1}{W}\sum_{j\in S_t}w_je_{i_t}V_{j}^T}$
       \State $\Delta V\leftarrow \Delta V - \frac{1}{N} \prns{e_{j_t}U_{i_t}^T-\frac{1}{W}\sum_{j\in S_t}w_je_{j}U_{i_t}^T}$
   \EndFor
   \Repeat
       \State $U' \leftarrow U+\eta\Delta U$, $V' \leftarrow V+\eta\Delta V$
       \State $f' \leftarrow L(U'V'^T) + \frac\lambda2\|U'\|_\text{F}^2 + \frac\lambda2\|V'\|_\text{F}^2$
       \State $\eta \leftarrow \beta_{\mathrm{dec}}\eta$
   \Until {$f' \leq f$}
   \State $U \leftarrow U'$, $V \leftarrow V'$
   \Until{$\frac{f-f'}{f'} \leq \tau$}
   \State {\bfseries output:} {$UV^T$}
\end{algorithmic}
\end{algorithm}

We initialize our algorithm using a technique from \citet{bhojanapalli2015dropping},
which only requires access to gradients of the objective of \eqref{eq-alg-alt}.
Using the SVD, we write $-\nabla L(0) = \tilde U \op{diag}(\tilde\sigma_1,\dots,\tilde\sigma_{\min(m,n)}) \tilde V^T$ and initialize
\[\begin{array}{rl}
U^0 &= {\gamma}^{-1/2} \op{diag}(\sqrt{\tilde\sigma_1},\dots,\sqrt{\tilde\sigma_{\tilde r}}) \tilde U_{:,\tilde r},\\
V^0 &= {\gamma}^{-1/2} \op{diag}(\sqrt{\tilde\sigma_1},\dots,\sqrt{\tilde\sigma_{\tilde r}}) \tilde V_{:,\tilde r},\\
\end{array}\]
where $\gamma = \|\nabla L(0) - (\nabla L(e_1 e_1^T) + \lambda e_1 e_1^T)\|_\text{F}$ and
$\tilde U_{:,\tilde r}$, $\tilde V_{:,\tilde r}$ denote the first $r$ columns of $\tilde U$, $\tilde V$.
We use an adaptive step size with a line search that guarantees descent.
Starting with a stepsize of $\eta=1$, the stepsize is repeatedly decreased by a factor
$\beta_{\mathrm{dec}}$ until the step produces a decrease in the objective.
We terminate the algorithm when the
decrease in the relative objective value
is smaller than the convergence tolerance $\tau$.

\section{The Dynamic Assortment Personalization Problem}\label{dynasssec}

We now return to the dynamic setting.
First, we define some notation that will be useful in the following discussion.
For $\theta\in\R n$, $w\in\R n$, $S\subset\braces{1,\dots,n}$, and $K\leq n$,
let
\begin{align*}
p_j(S;\theta)&=\frac{e^{\theta_{j}}}{1+\sum_{j'\in S}e^{\theta_{j'}}},~&&\text{the MNL choice probability under parameters $\theta$,}\\
F(S;w,\theta)&={\sum_{j\in S}p_j(S;\theta)w_{j}},~&&\text{the expected revenue of assortment $S$ under $\theta$, and}\\
S^\star(w,\theta;K)&={\operatornamewithlimits{argmax}_{\abs{S}\leq K}F(S;w,\theta)},~&&\text{the {set of optimal assortments} under $\theta$ for revenues $w$.}
\end{align*}

\subsection{Structure-ignorant algorithms}
An algorithm is \emph{structure-ignorant} if it ignores any potential structure that connect
the different contexts (rows of $\Theta^\star$). Therefore, a structure-ignorant algorithm
is one that runs separate, independent algorithms for each context. Formally,
we make the following definition.

\begin{definition}
An algorithm $\pi$ for the dynamic assortment personalization problem is
\emph{structure-ignorant} if, under $\pi$, the variable
$S_t$ is measurable with respect to only the historical
data from type $i_t$,
$\{i_{t'},\,j_{t'},\,S_{t'}:t'<t,i_{t'}=i_t\}$, and
the variables $U_1,\dots,U_t$.
\end{definition}

An algorithm is said
to be consistent if it has sublinear regret in $T$ over all problem instances.
\begin{definition}
Fix $(m,n,W)$. An algorithm $\pi$ is said to be \emph{consistent} if for any $\mu^\star,\,\Theta^\star$, and $a>0$, we have
$\op{Regret}(T;\pi)=o(T^a)$ over all problem instances $(m,n,W,\mu^\star,\Theta^\star)$.
\end{definition}
For brevity, we usually omit the subscript and abuse notation in referring to a family or sequence of algorithms simply as one algorithm $\pi$.

If we run separate algorithms for each context then, in each context, we are solving
the classic (non-contextual) dynamic assortment planning problem, precisely as studied by \citet{rusmevichientong2010,saure2013}.
Making use of Theorem 1 of \citet{saure2013} we establish the following:
\begin{theorem}\label{regcor}
Fix $\nu\in(0,1)$, $w\geq0$, $\rho\geq1$.
Given any family (indexed by revealed problem parameters $m,n,W$) of structure-ignorant consistent algorithms $\pi$,
we have
$$\op{Regret}(T;\pi)=\Omega(\min(m,n)\max(m,n)\log T)$$
over all times $T$ and problem instances $(m,n,W,\mu^\star,\Theta^\star)$ such that
\begin{enumerate}
\item $T$ grows superlinearly in the number of types $m$: for some $\epsilon>0$,
$T = \Omega(m^{1 + \epsilon})$.
\item For every type $i=1,\dots,m$:
\begin{enumerate}
\item Profit is bounded: \(\|W_i\|_\infty \leq w.\)
\item The type appears often enough and not too often:
\(
1/\rho\leq m\mu_i^*\leq \rho.
\)
\item The number of \emph{potentially optimal items} grows linearly in $n$:
\[
\abs{\braces{j\notin S^\star(W_i,\Theta^\star_i;K):~
\exists\theta\in\R n,\,j\in S^\star(W_i,\theta;K),\,\theta_{j'}=\Theta^\star_{ij}~
\forall j'\in S^\star(W_i,\Theta^\star_i;K)}}\geq\nu n.
\]
\end{enumerate}
\end{enumerate}
\end{theorem}

\subsubsection{\beditb Proof sketch for Theorem~\ref{regcor}}\beditb
To prove Theorem~\ref{regcor}, we argue as follows. Let $r(i;S)=\frac{\sum_{j\in S}\exp{\Theta^\star_{ij}}W_{ij}}{1+\sum_{j\in S}\exp{\Theta^\star_{ij}}}$ so that $R_t=\max_{\abs{S}\leq K}r(i_t;S)-r(i_t;S_t)$ is the expected instantaneous regret at time $t$, averaged over $j_t$
Let $T_i=\sum_{t=1}^T\indic{i_t=i}$ be the number of times type $i$ is encountered. We use a conditioning argument, a concentration bound on $T_i$, and the fact that $T=\Omega(m^{1+\epsilon})$ to argue that
$$\E \op{Regret}_T(\pi)=\E\left[\sum_{t=1}^T R_t\right]\geq\frac12 \left(\sum_{i=1}^m \E\left[\sum_{t:~ i_t = i} R_t \,\middle\vert\, T_i \geq \frac{T \mu_i}{2} \right]\right).$$
This bound shows that the regret is at least half the sum of regrets incurred in each context seen at least $T\mu_i/2$ times. We invoke \citet[Theorem 1]{saure2013} to provide a lower bound on the right hand side for each context when the algorithm plays each context independently.

\subsubsection{Discussion of Theorem~\ref{regcor}}
\eeditb
Note that $mn=\min(m,n)\max(m,n)$.
We rewrite the product in this way to make the comparison with our
structure-aware algorithm more clear:
it depends on $\max(m,n)$, but replaces the $\min(m,n)$ term by the rank $r$
of the parameter matrix $\Theta$.

Theorem~\ref{regcor} asserts that structure-ignorant algorithms for the dynamic assortment personalization
problem have regret that grows linearly in the number of item-type combinations
and logarithmically in the time horizon.
That is, ignoring structure incurs \emph{enormous} regret.
In the sequel, we improve on this regret bound by developing a structure-aware algorithm.

\subsection{A Structure-Aware Algorithm}\label{s-two-phase}

In Section \ref{s-model}, we argued that imposing structure is crucial for
learning the preferences of a very heterogeneous population and
proposed an estimator that can leverage structure to learn the {LRCMNL} model
in sublinear time.
In this section, we use this estimator to develop a structure-aware algorithm
that can achieve regret sublinear in problem size $mn$.

In Algorithm \ref{alg:nucnormpolicy}, we define an algorithm for
dynamic assortment personalization with tuning parameters $C$ and $\lambda$.
We refer to this algorithm as $\pi_\text{nuc-norm}$.

\begin{algorithm}[t!]
   \caption{Dynamic Assortment Personalization
   ($\pi_\text{nuc-norm}(C, \lambda)$) \label{alg:nucnormpolicy}}
\begin{algorithmic}\SingleSpacedXI
\State {\bf input:} $C$, $\lambda$
\State Initialize set of randomized observations $\mathcal O = \varnothing$.
\vspace{1em}
\For{$t=1,2,\dots$}
\If{$|\mathcal O| \leq C r(m+n) \log(t)$}
\State \textit{Explore:}
  \State \hspace{\algorithmicindent} observe customer type $i_t$,
  \State \hspace{\algorithmicindent} choose $S_t$ uniformly at random from all subsets of size $K$,
  \State \hspace{\algorithmicindent} observe customer product choice $j_t$, and
  \State \hspace{\algorithmicindent} update the set of randomized observations $\mathcal O \leftarrow \mathcal O \cup (i_t, j_t, S_t)$.
\State \textit{Estimate:} let
\[
L(\Theta)=\frac1{|\mathcal O|}\sum_{(i,j,S) \in \mathcal O} \log\prns{\prns{1+\sum_{j'\in S}e^{\Theta_{ij'}}}\prns{\left\{\ba{lc}1 \quad\quad& j=0\\e^{\Theta_{ij}} \quad\quad& \text{otherwise}\ea\right.}^{-1}},
\]
\State \phantom{\textit{Estimate:}} and solve
\begin{equation*}
\ba{rl}
\widehat\Theta\in\operatornamewithlimits{argmin} \quad& L(\Theta) + \lambda\|\Theta\|_* \\
\text{s.t.} \quad& \|\Theta\|_\infty\leq\alpha/\sqrt{mn}.
\ea
\end{equation*}
\Else
\State \textit{Exploit:} observe customer type $i_t$ and choose $S_t\in S^\star(W_{i_t},\widehat\Theta_{i_t};K)$.
\EndIf
\EndFor
\vspace{1em}
\State {\bf output:} $S_1, S_2, \ldots$
\end{algorithmic}
\end{algorithm}

Every step of Algorithm~\ref{alg:nucnormpolicy} is computationally tractable,
including the computation of the optimal assortment
$S^\star(W_{i_t},\widehat\Theta_{i_t};K)$ \citep{rusmevichientong2010} and
of $\widehat \Theta$ (Algorithm~\ref{alg:fgd}).
{
The optimal assortments $S^\star(W_{i},\widehat\Theta_{i};K)$ for each type $i$
need only be recomputed when the parameter estimate $\widehat\Theta$ changes.
These changes happen only on exploration steps, which occur a vanishing fraction of the time
as $t$ increases.
Hence we need only compute optimal assortments a vanishing fraction of the time.
}

Next we show that this achieves regret that is orders of magnitude smaller
than a structure-ignorant algorithm when structure is present.
Let
$$\delta(w,\theta;K)=\max_{\abs{S}\leq K}F(S(w,\theta;K);w,\theta)-\max_{S\notin S^\star(w,\theta;K) : \abs{S}\leq K}F(S;w,\theta)$$
be the gap in revenue between the optimal assortment and any suboptimal assortment under MNL choice with parameter vector $\theta$.

\begin{theorem}\label{regbound1}\beditb
Let
$r$ be such that $r\geq\op{rank}(\Theta^*)$,
$\rho\geq1$ be such that $1/\rho\leq m\mu_i\leq \rho$ $\forall i=1,\dots,m$,
$\alpha$ be such that $\magd{\Theta^\star}_\infty\leq\alpha/\sqrt{mn}$, $\delta$ be such that $\min_{1\leq i\leq m}\delta(W_i,\,\Theta^\star_i;\,K)\geq\delta$, and $\omega$ be such that $\magd{W}_\infty\leq\omega$.
Choose as algorithm parameters 
\begin{equation}\label{banditalgoparams}
C=4194304K^6\rho^3\omega^2 \alpha^2\exp{(16\alpha)} / \delta^2, \quad \lambda = 8\sqrt{\frac{\rho K}{Crmn}}.
\end{equation}
Then the regret of $\pi_\text{nuc-norm}(C,\,\lambda)$
satisfies
\begin{align*}
\op{Regret}\prns{T;\pi_\text{nuc-norm}(C,\,\lambda)}&\leq ((Cr(m+n)+3)\log(T)+1)\omega
\\&=O\prns{r\max(m,n)\log(T)}
\end{align*}
for all $T\leq (m+n)^{\frac{mn}{C(m+n)}r}$.\eeditb
\end{theorem}

\beditb
\subsubsection{Proof sketch for Theorem~\ref{regbound1}}
To prove Theorem~\ref{regbound1}, we argue as follows.
We first establish that expected revenue is smooth with respect to the parameter $\theta$:
$\abs{F(S;w,\theta)-F(S;w,\theta')}\leq\frac14\magd{w}_\infty K^{3/2}\magd{\theta-\theta'}_2$.
Therefore, for any two different parameter vectors $\theta$ and $\theta'$
together with corresponding optimal sets $S\in S^\star(w,\theta;K)$ and $S'\in S^\star(w,\theta';K)$,
the expected revenue under parameter $\theta$ cannot change much when we replace the set $S$ by $S'$:
$$
F(S';w,\theta)\geq F(S ;w,\theta )-\frac12\magd{w}_\infty K^{3/2}\magd{\theta-\theta'}_2.
$$
Hence if $\magd{\theta-\theta'}_2$ is small enough,
the optimal assortments for each are the same:
$$
\prns{\magd{\theta-\theta'}_2\leq2\delta(w,\theta;K)K^{-3/2}/\magd{w}_\infty}\implies \prns{S^\star(w,\theta';K)=S^\star(w,\theta ;K)}.
$$
We combine this result with Theorem~\ref{mainthm} (choosing $\tau=\log(t)/\log(m+n)$)
to argue that the average instantaneous regret at any time $t$
when Algorithm~\ref{alg:nucnormpolicy} chooses to exploit must be bounded as
$$\E R_t\leq3\omega/t.$$
To complete the proof, note that Algorithm~\ref{alg:nucnormpolicy}
chooses to explore no more than $O(r\max(m,n)\log(T))$ times.

\subsubsection{Discussion of Theorem~\ref{regbound1}}

\beditb
{
Notice that the time horizon $T$ appears {in} Theorem~\ref{regbound1},
but does not appear in Algorithm~\ref{alg:nucnormpolicy}.
In particular, the horizon $T$ need not be known in advance
to run Algorithm~\ref{alg:nucnormpolicy}, and
for any time $T\leq (m+n)^{\frac{mn}{C(m+n)}r}$
Theorem~\ref{regbound1} provides
a valid regret bound {on our horizon-independent algorithm}.
Specifically, this restriction applies in the high-dimensional setting
of interest, with
no more observations than the number of item-type combinations
($T=O(mn)$). If Theorem~\ref{regbound1}'s restriction were violated,
which is unrealistic in the
high-dimensional setting, it would mean
that there is enough time to leisurely learn each user's preferences
completely separately as even a structure ignorant algorithm
can achieve vanishing time-average suboptimality:
$mn\log(T)/T\to0$ as $m,n\to\infty$. Thus, Theorem~\ref{regbound1}
exactly captures the regime of interest.

\begin{table}[t!]\centering\SingleSpacedXI
\beditb\caption{{Regret relative to horizon $T$ in different horizon regimes with bounded rank, $r=O(1)$.
}}\label{regrettable}
\begin{tabular}{cccc}
\toprule
    \multirow{3}{*}{Regret}& Uninformative   & High-dimension &Low-dimension\\\cmidrule{2-4}
    &           \multirow{2}{*}{$T=\tilde O(\max(m,n))$} & $T=\tilde\omega(\max(m,n))$ & \multirow{2}{*}{$T=\tilde \Omega(mn)$}\\
    &&$T=\tilde o(mn)$&\\\midrule
$r\max(m,n)\log(T)$ & Linear & Sublinear & Sublinear\\
$mn\log(T)$ &  Linear & Linear & Sublinear\\\bottomrule
\end{tabular}\eeditb
\end{table}

Comparing to Theorem~\ref{regcor},
Theorem~\ref{regbound1} shows that leveraging
structure appropriately can lead to
regret that is an order of magnitude smaller
than that of structure-ignorant algorithms.
Specifically, we find that the ratio of the regret of \emph{any} structure-ignorant algorithm to that of our algorithm grows at least as fast as $\Omega\prns{\frac{\min(m,n)}r}$.
Since $\min(m,n) \geq r$, our algorithm performs at least as well as \emph{any} structure-ignorant algorithm.
And whenever the rank of $\Theta^\star$ grows more slowly than its side dimension,
$r=o(\min(m,n))$, our algorithm performs \emph{strictly} better.
The improvement is largest when the rank $r$ is constant:
in this case, the regret ratio grows \emph{linearly} in the side dimension $\min(m,n)$.
Whenever $r=o(\min(m,n))$, because our regret grows more slowly, the difference between the regret of any structure-ignorant algorithm to that of our algorithm is $\Omega(mn\log(T))$. {In other words, the additional revenue that our algorithm generates relative to any structure-ignorant algorithm grows at such a rate, highlight the revenue impact of our algorithm.}

In order to understand the different regret regimes, let us consider a setting with equal side dimension, $m=n$, and bounded rank, $r=O(1)$.
If $T=O(m)$ then we can only interact with each user at most a constant number of times.
Therefore it is impossible to consistently learn
the choice model, since the number of observations is of the same order of the number of parameters.
Hence the regret of any algorithm must be linear in $T$.
We must have more observations ($T=\omega(m)$) to learn the choice model consistently and just slightly more than that ($T/\log(T)=\omega(m)$) to ensure sublinear regret.
If the horizon $T$ is slightly longer than the number of users, $T/\log(T)=\omega(m)$,
our algorithm, which incurs regret $m\log(T)$ in this case, will achieve sublinear regret in $T$.
However, in the same setting, as long as $T/\log(T)=o(m)$,
a structure ignorant algorithm incurs linear regret in $T$ (by Theorem~\ref{regcor}).
If the horizon $T$ is much larger than the number of user-item combinations, $T/\log(T)=\Omega(m^2)$,
than even algorithms that ignore structure can achieve sublinear regret:
we have enough observations to learn each user's preferences independently.
Hence both our algorithm and any efficient structure ignorant algorithm
(e.g., using the method of \cite{saure2013} for each user separately) achieve sublinear regret.

We summarize the above observations in Table~\ref{regrettable} in the more general unequal side dimension case.
In the table we abbreviate $T/\log(T)=\Omega(mn)$ as $T=\tilde\Omega(mn)$ and similarly for $\tilde o,\tilde\omega$.
The table shows the three informational regimes:
an uninformative regime where there can be no hope of linear regret,
a high-dimensional regime where algorithms must use cross-user information and latent structure
to achieve sublinear regret,
and a low-dimensional regime where cross-user information is not required achieve sublinear regret.




}
\eeditb
{
{Note that in Theorem~\ref{regbound1},
that the choice of parameters $C,\lambda$ for
Algorithm~\ref{alg:nucnormpolicy} depend on
the lower bound, $\delta$, on the revenue gap between the optimal
assortment and any suboptimal assortment.}
This dependence agrees with previous work on dynamic assortment planning
\citep{rusmevichientong2010,saure2013}, which also proposes algorithms
and regret analyses that require prior knowledge of this gap.
Removing this dependence in our Algorithm~\ref{alg:nucnormpolicy}
is an important problem for future research.
Recently, \citet{agrawal2017mnl} addressed this issue in the classic,
non-contextual dynamic assortment planning problem,
developing a new algorithm that does not depend
on the optimality gap.
}







\section{Experimental Results}\label{numerics}

\begin{figure}
\centering
\caption{RMSE of estimators for $\Theta^\star$ with $m=n=100,150,200,\dots,500,750,\dots,4000$ and $r=2,\,25$.}\label{thefigure}
\includegraphics[width=0.4375\textwidth]{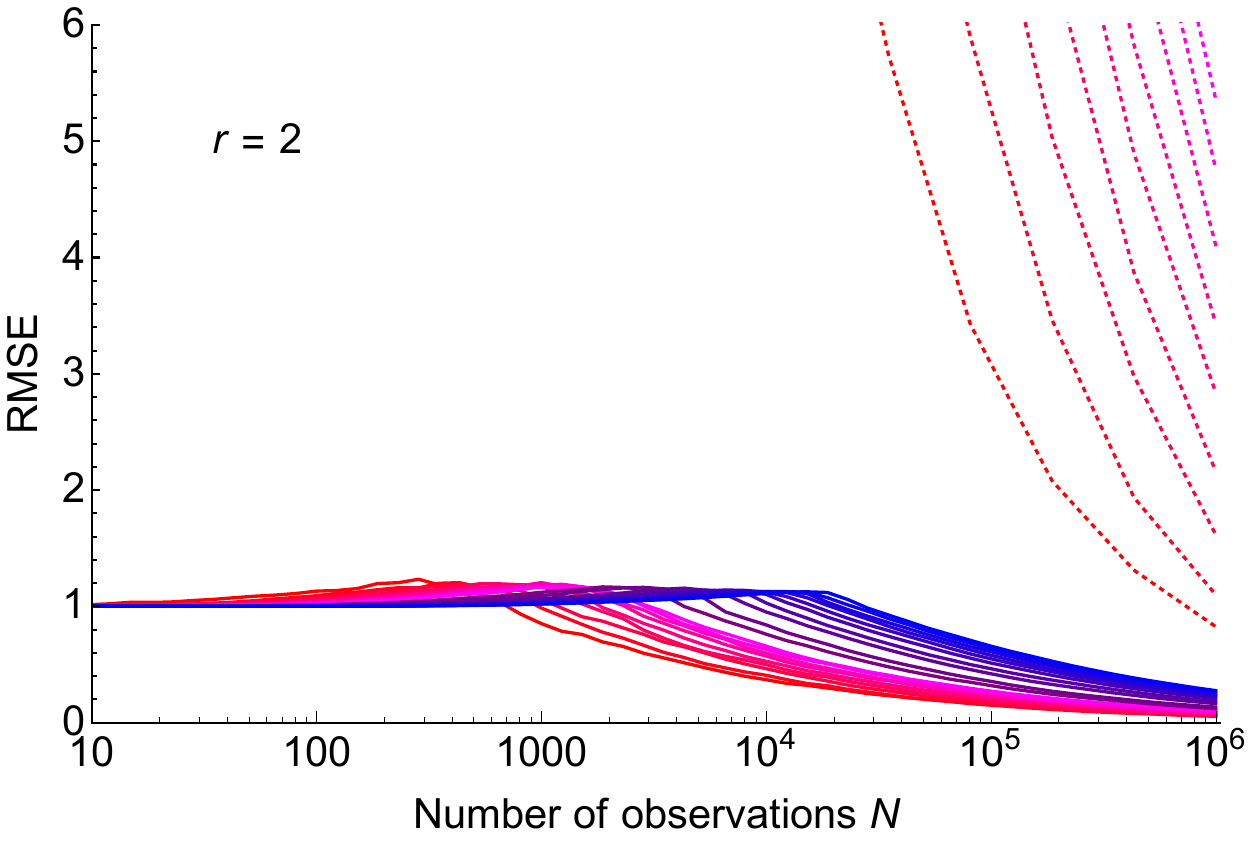}
\includegraphics[width=0.4375\textwidth]{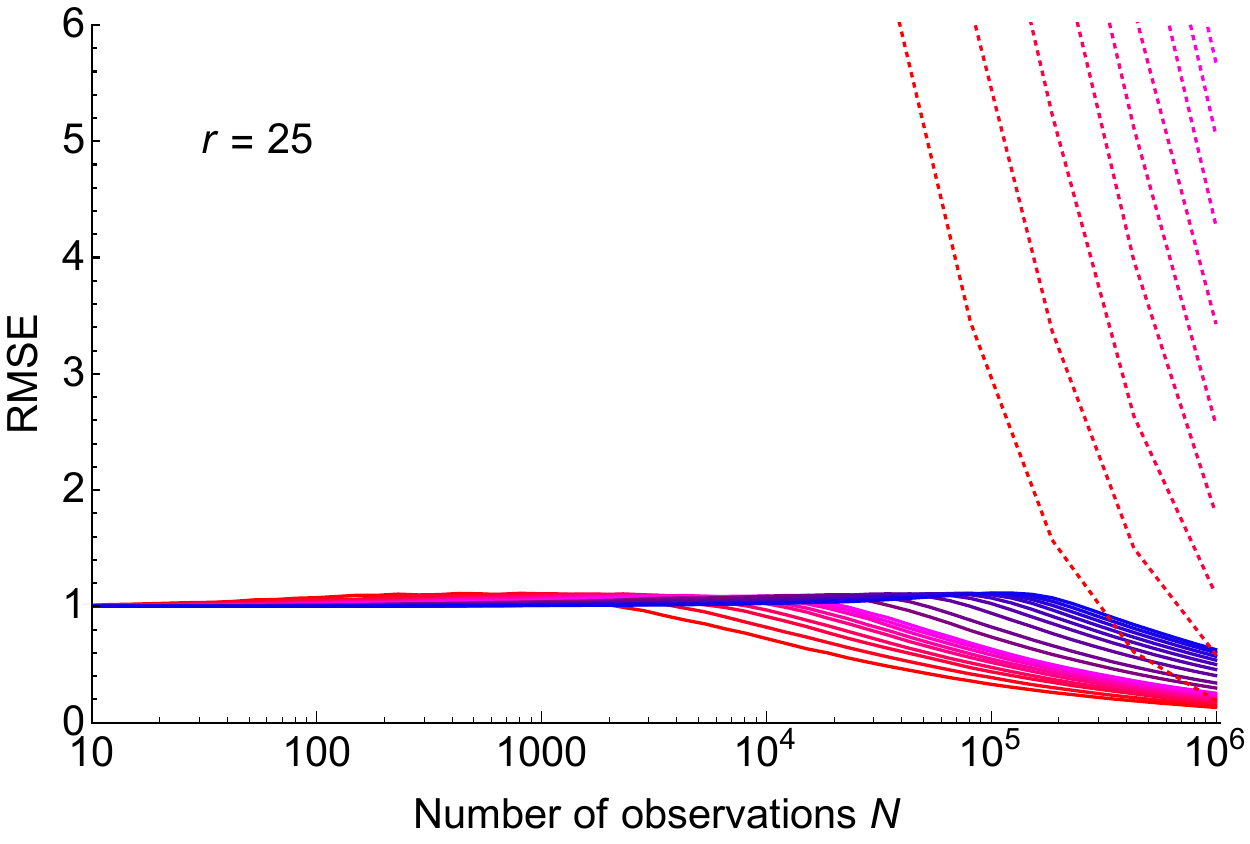}
\raisebox{1.25em}{\includegraphics[width=0.125\textwidth]{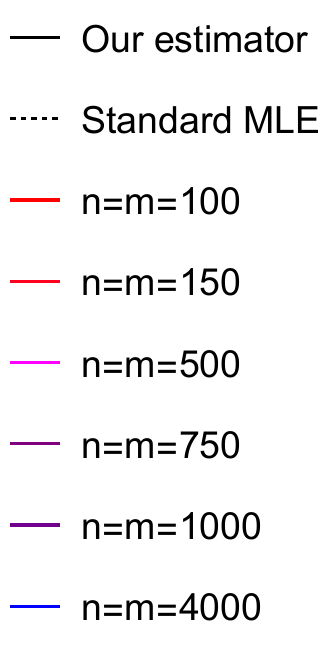}}
\end{figure}

In this section we demonstrate numerically the importance of structure and the power of our approach.

\subsection{The Static Estimation Problem}

First, we focus on the static estimation problem. We compare our estimate $\widehat\Theta$ with the standard maximum
likelihood estimate $\widehat\Theta^\text{MLE}$ that solves
\begin{equation}
\label{eq-mle}
\ba{rl}
\mbox{minimize} & L(\Theta).
\ea
\end{equation}

Note that, since it imposes no structure on the whole matrix $\Theta$,
problem \eqref{eq-mle} decomposes into $m$ subproblems for each type (row of $\widehat\Theta^\text{MLE}$),
each solving a separate MNL MLE in $n$ variables.
In our experiments, we use Newton's method as implemented by \texttt{Optim.jl}
to solve each subproblem.

To generate $\Theta^\star$, we fix $m,\,n,\,r$, let $\Theta_0$ be an $m\times n$
matrix composed of independent draws from a standard normal, take its SVD
$\Theta_0=U\op{diag}\prns{\sigma_1,\sigma_2,\dots} V^T$,
truncate it past the top $r$ components $\Theta_1=U\op{diag}\prns{\sigma_1,\dots,\sigma_r,0,\dots} V^T$,
and renormalize to achieve unit sample standard deviation to get $\Theta^\star$, \ie, $\Theta^\star=\Theta_1/\op{std}(\op{vec}(\Theta_1))$.
To generate the choice data, we let $i_t$ be drawn uniformly at random
from $\{1,\dots,m\}$, $S_t$ be drawn uniformly at random from all subsets
of size $10$, and $j_t$ be chosen according to
\eqref{themodel} with parameter $\Theta^\star$.

For our estimator we use Algorithm \ref{alg:fgd}
with $\tilde r=2r$, $\lambda=\frac18\sqrt{\frac{Kd\log d}{mnN}}$,
$\beta_{\mathrm{dec}} = 0.8$, and $\tau=10^{-10}$.
This regularizing coefficient scales with $m$, $n$, $d$, $N$, and $K$ as
suggested by Theorem \ref{mainthm}, but we find the algorithm performs
better in practice when we use a smaller constant than that suggested
by the theorem.

We plot the results in Figure \ref{thefigure}, where error is measured in
root mean squared error (RMSE)
$$
\op{RMSE}(\Theta)
=\sqrt{\op{Avg}\prns{\{(\Theta_{ij}-\Theta^\star_{ij})^2\}_{i,j}}}
=\frac{1}{\sqrt{mn}}\magd{\Theta-\Theta^\star}_\text{F}.
$$
The results show the advantage in efficient use of the data offered by our approach.
The results also show that, relative to MLE,
the advantage is greatest when the underlying rank $r$ is small and the number of parameters $m\times n$ is large, but that we maintain a significant advantage even for moderate $r$ and $m\times n$. For large numbers of parameters ($m=n\geq 750$), the RMSE of MLE is very large and does not appear in the plots.
Only in the case of greatest rank ($r=25$), smallest number of parameters ($m=n=100$), and greatest number of observations ($T=10^6$) does MLE appear to somewhat catch up with our estimator.

\begin{figure}
\centering\caption{RMSE of our estimators for $\Theta^\star$ by observation per row with $r=2$.}\label{thefigure2}
\includegraphics[width=0.9\textwidth]{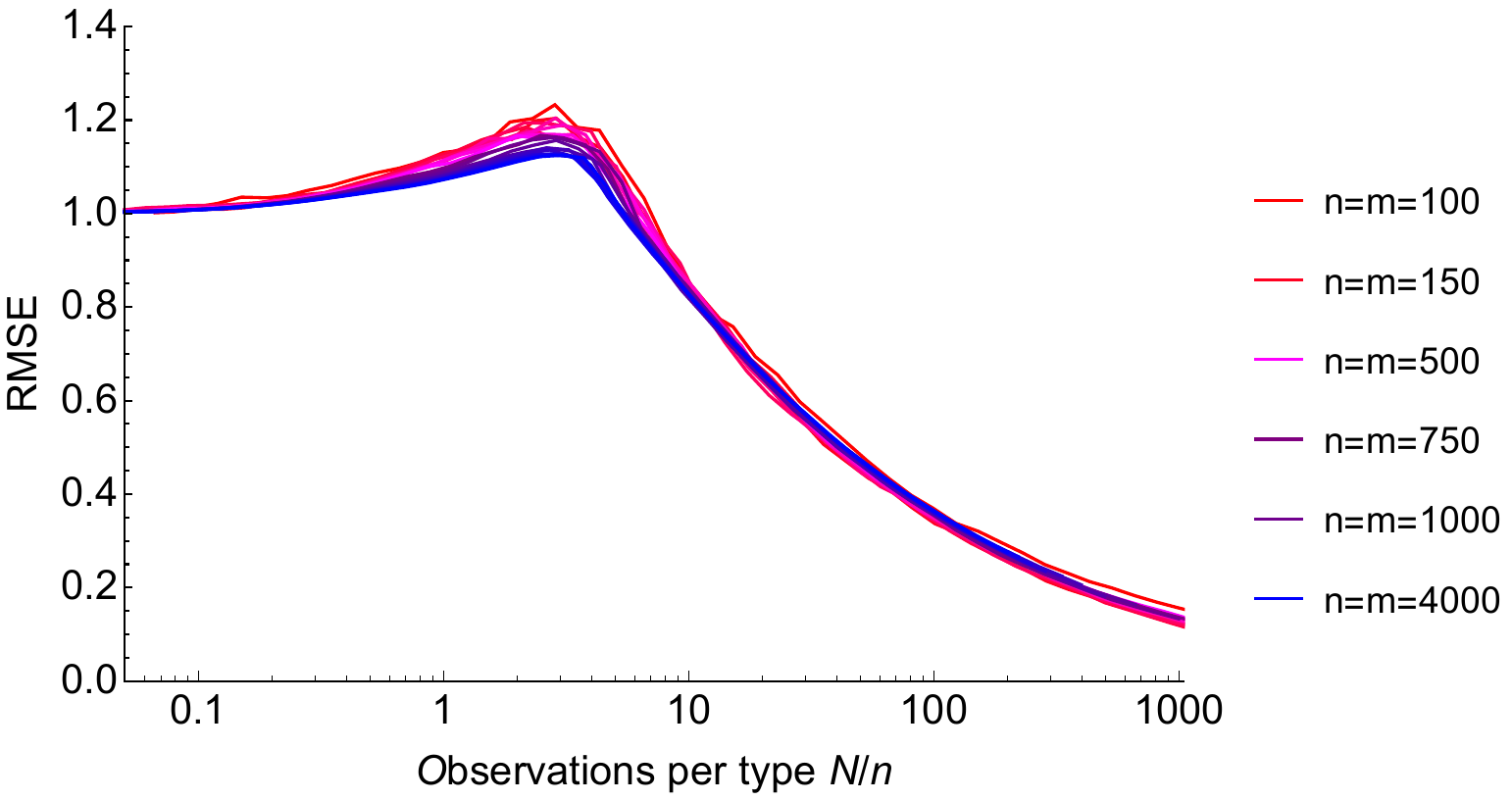}
\end{figure}

In Figure \ref{thefigure2}, we plot the RMSE of our estimator
against the number of observations per type (or item) $N/d$ for a square problem with $d=m=n$.
We see nearly the same error curve traced out as we vary the problem size $d$.
This scaling shows that our estimator is able to leverage the low-rank
assumption and require the same number of choice observations per type to achieve the same RMSE regardless of problem size.
{Thus, even in high dimensions, if customer behavior is dictated
by a bounded number of (latent) factors,
then our approach need only rely on a
{small}
number of observations per type.}


\begin{figure}[t!]
\centering
\caption{Regret for Structure-Aware, Structure-Ignorant, and Context-Ignorant Algorithms (note horizontal log scale)}\label{regfig}
\subfloat[{$m=n=100$, $mn=10,00$}]{
\includegraphics[width=0.475\textwidth]{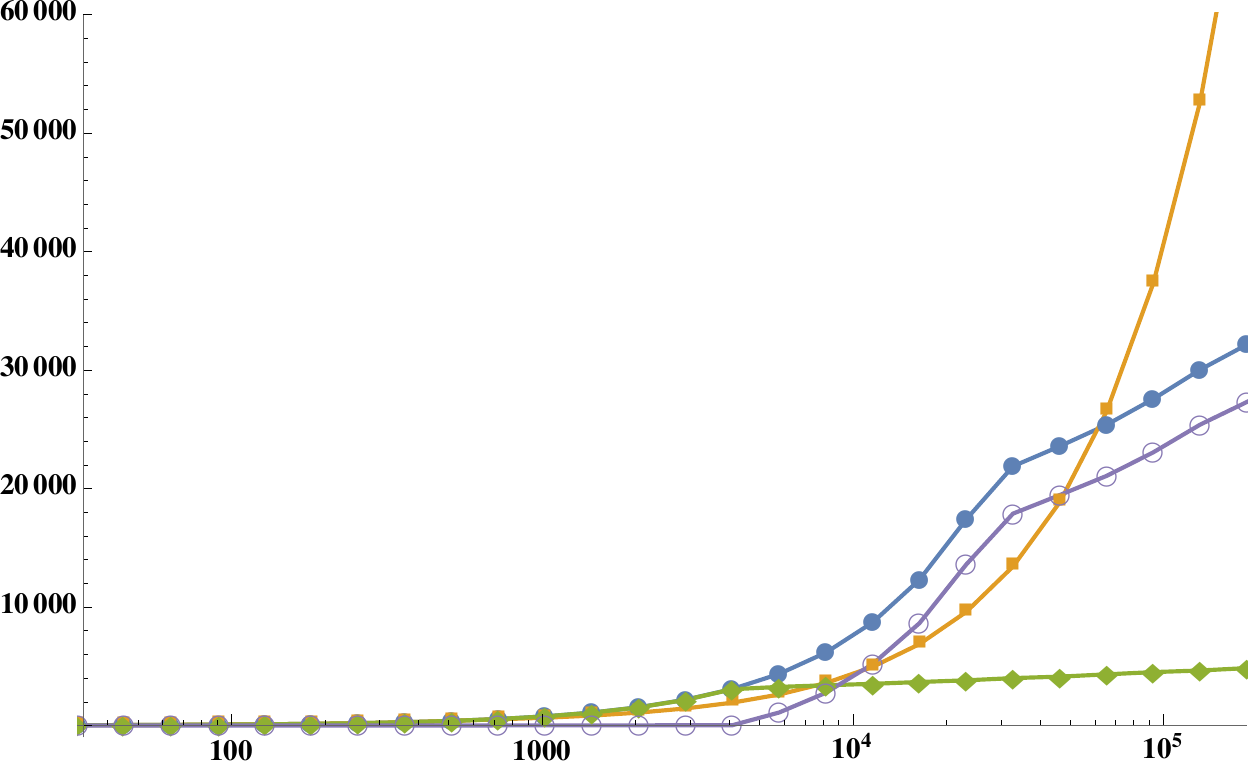}
}
\subfloat[{$m=n=200$, $mn=40,000$}]{
\includegraphics[width=0.475\textwidth]{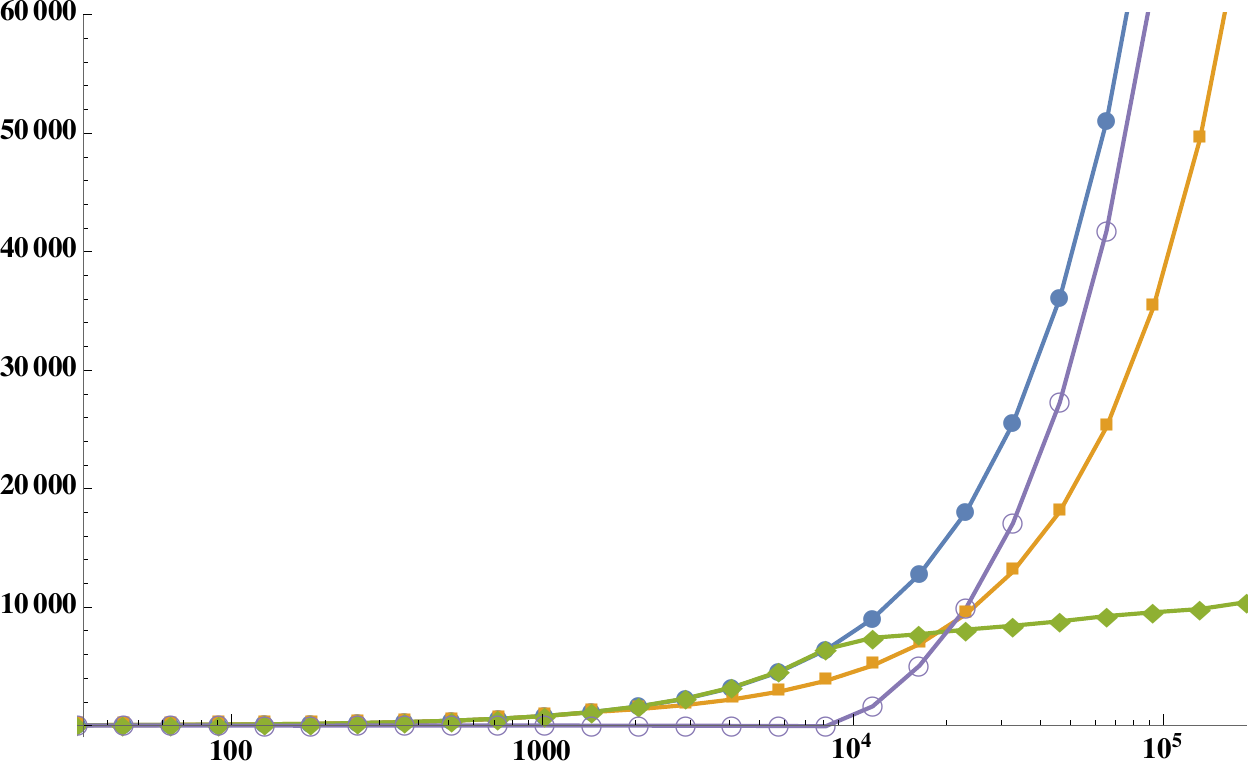}
}
\\
\subfloat[{$m=n=300$, $mn=90,000$}]{
\includegraphics[width=0.475\textwidth]{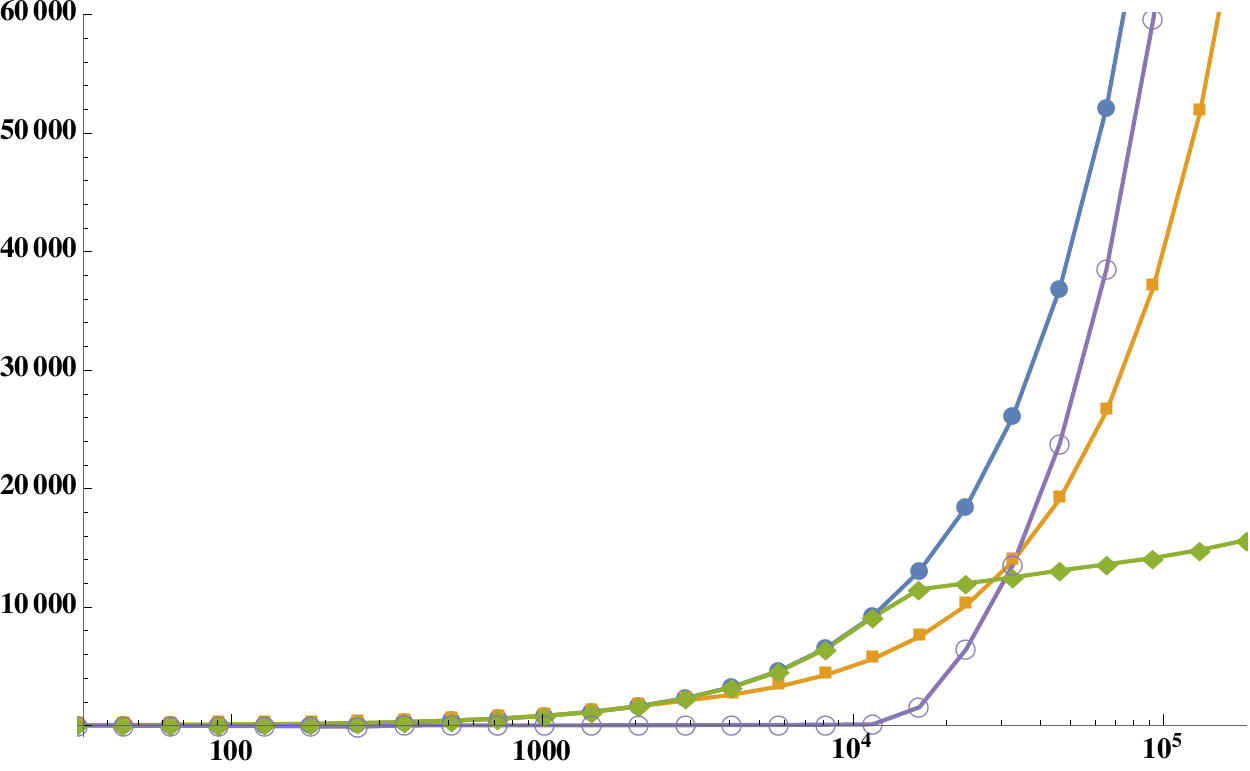}
}
\subfloat[{$m=n=400$, $mn=160,000$}]{
\includegraphics[width=0.475\textwidth]{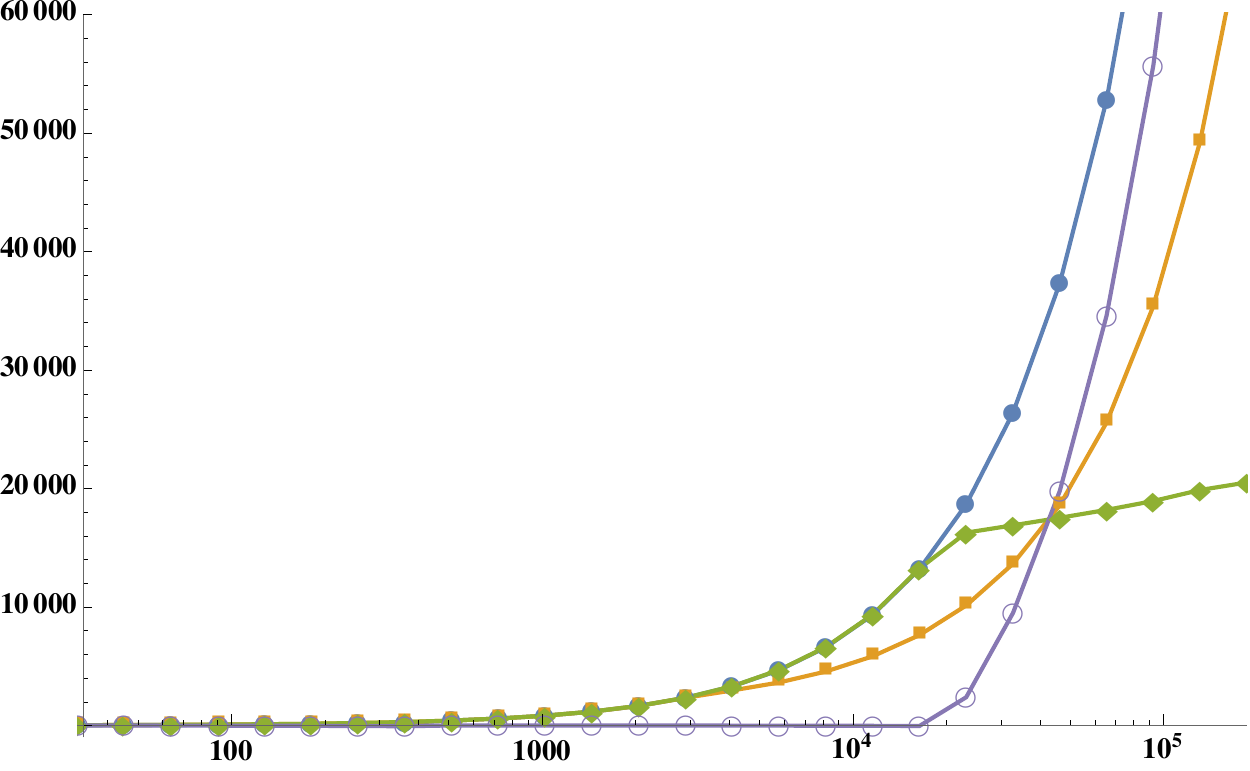}
}
\\
\raisebox{3em}{\includegraphics[width=0.9\textwidth]{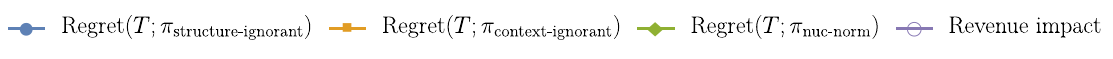}}
\end{figure}

\subsection{The Dynamic Assortment Personalization Problem}

Next, we turn our attention to the dynamic assortment personalization problem.
We compare our algorithm $\pi_\text{nuc-norm}$ 
to two alternatives. \beditb One alternative ($\pi_\text{structure-ignorant}$) is the structure-ignorant algorithm in which we apply Algorithm 1 of \citet{saure2013} to each type separately.
Recognizing that one cannot learn a huge set of parameters from very few observations,
the second alternative ($\pi_\text{context-ignorant}$) tries to fit a single MNL model to the whole population, applying only one replicate Algorithm 1 of \citet{saure2013} to all types simultaneously.
We generate the problem exactly as in the previous section and run each of these algorithms on problem dimensions varying from 10,000 to 160,000 and for $r=2$. We plot the regret of each of these algorithms over time in Figure \ref{regfig}. We also include the regret of $\pi_\text{structure-ignorant}$ relative to our $\pi_\text{nuc-norm}$ (``Revenue impact''), which is just the difference of their regrets relative to the optimal policy since this baseline will cancel. This quantifies the revenue impact of our algorithm in terms of the additional revenue we can generate using our structure-aware algorithm.\eeditb

Since the plots have a logarithmic horizontal axis, regret that is logarithmic in $T$ appears as a line in the figure whereas regret that is linear in $T$ appears as an exponential function.
These plots reveal several interesting features of the algorithms.

We see that the structure-ignorant algorithm, which we know has logarithmic regret asymptotically, only achieves logarithmic regret for the smallest of the problems ($m=n=100$). For larger problem sizes, regret appears to be linear for all horizons $T$ shown.
The transition from linear to logarithmic regret is not visible because the problems are so large that the transition occurs for extremely large $T$: on the scale of this plot, the $mn$ term overwhelms the $\log(T)$ term.

In all of these larger problems, the context-ignorant algorithm performs better than the structure-ignorant algorithm. Both have linear regret in this parameter regime. In fact, the context-ignorant algorithm uses a misspecified model, and so has asymptotically linear regret. Hence for very large $T$ (not shown on this plot) it will be overtaken by the structure-ignorant algorithm, whose regret is asymptotically logarithmic. The success of the misspecified context-ignorant algorithm holds an important lesson: when time is limited, it is more effective to use a misspecified model with few parameters than a well specified model with many parameters.

On the other hand, our structure-aware algorithm exhibits logarithmic regret in each and every case, even at this relatively short time scale. The algorithm $\pi_\text{nuc-norm}$, as promised by Theorem \ref{regbound1}, has logarithmic regret that does not explode astronomically with $m,n$. \beditb Correspondingly, it achieves significant revenue impact relative to the structure-ignorant algorithm as shown in Figure~\ref{regfig}.\eeditb

\section{Conclusion}

To manage revenue, many retailers must solve a dynamic assortment personalization problem: they must learn customers preferences in real time, at scale,
from customers' choices from among the items on offer;
and they must quickly use this information to present revenue-maximizing assortments.
This paper explores a structural approach to enable large scale dynamic assortment personalization. We proposed algorithms using structural (low rank) priors to learn and exploit customer preferences.
We presented theoretical and numerical evidence that these algorithms improve on the state of the art by orders of magnitude, and achieve performance suitable for use in practice.


\ACKNOWLEDGMENT{
The authors thank several anonymous reviewers for their guidance in
improving this manuscript.
NK was supported by the National Science Foundation under Grant No. 1656996.
MU was supported by DARPA Award FA8750-17-2-0101
and by a fellowship awarded by the Center for the Mathematics of Information at Caltech.}

\bibliographystyle{informs2014}
\bibliography{DynAssortHiDim}


\clearpage
\noindent{\Large Appendices}

\beditb
\section{Dynamic Assortment Planning with a Heterogeneous Population}

In this section we consider a \emph{non-contextual} dynamic assortment planning problem where the target population is very heterogeneous.
This is a departure from our focus on personalization, but theoretically our results can be extended to this additional setting that may be of independent interest.

This setting is appropriate when there is no potential for user-level personalization, such as in a single brick-and-mortar store,
but where the consumer population is very heterogeneous and cannot be well described by only a single, or even a few, nominal preference vectors and where this heterogeneity can be explained by an post-purchase observable user type.
This setting is similar to that considered in
\cite{rusmevichientong2010} and \cite{saure2013} in that it is non-contextual, but in both those works
the choice distribution is assumed to be MNL (or, generally, any random utility model with a single nominal
preference vector, from which deviations are made at random in a homogeneous manner).
Here we allow for a potentially heterogeneous population with choice governed by the {LRCMNL} model.

The problem proceeds as follows, in somewhat different order than the personalization problem presented in Section \ref{mainprobsec}. At each $t=1,2,\dots$:
\begin{enumerate}
\item we choose any $S_t\subset\{1,\dots,n\}$ with $\abs{S_t}\leq K$,
\item a type $i_t$ is drawn at random from $\{1,\dots,m\}$ with probability proportional to weights $\mu^\star_i$,
\item an item $j_t$ is drawn at random from $\{0,1,\dots,n\}$ with probability proportional to weights
$$
\op{weight}(j)=\left\{\ba{lc}
1&\quad\quad{j=0}\\
0&\quad\quad{j\neq0,\,j\notin S_t}\\
\exp(\Theta^\star_{i_tj})&\quad\quad{j\neq0,\,j\in S_t}
\ea\right.
$$
\item if $j_t=0$ we get reward $r_t=0$ and otherwise we get reward $r_t=W_{i_tj_t}$.
\end{enumerate}
Unlike before, we have to select $S_t$ \emph{before} observing $i_t$. Therefore, we cannot personalize.

Let us re-define
\begin{align*}
F(S;W,\Theta,\mu)&=\sum_{i=1}^m\mu_i\frac{\sum_{j\in S}e^{\Theta_{ij}}W_{ij}}{1+\sum_{j\in S}e^{\Theta_{ij}}},~&&\text{the expected revenue of $S$ under $\Theta$, and}\\
S^\star(W,\Theta,\mu;K)&=
{\arg\max_{\abs{S}\leq K}F(S;W,\Theta,\mu)},~&&\text{the optimal assortment under $\Theta$.}
\end{align*}
Note that $S^\star(W,\Theta,\mu;K)$ is not efficiently computable; however, an
efficient approximation scheme, which searches over revenue-ordered assortments,
is proposed in \cite{rusmevichientong2014} and shown to work well.

We adapt our nuclear-norm-regularized algorithm to this case as shown in Algorithm \ref{alg:nucnormplan}. We refer to this algorithm as $\pi_\text{nuc-norm-plan}$.
The only difference between Algorithm \ref{alg:nucnormplan} and Algorithm~\ref{alg:nucnormpolicy} is that to exploit
our knowledge of $\widehat \Theta$, we choose any set in $S^\star(W,\widehat\Theta,\hat\mu;K)$
rather than observing $i_t$ and choosing a set in $S^\star(W_{i_t},\widehat\Theta_{i_t};K)$.


\begin{algorithm}[t!]
\SingleSpacedXI
   \caption{Dynamic Assortment Planning ($\pi_\text{nuc-norm-plan}(C,\,\lambda)$)\label{alg:nucnormplan}}
\begin{algorithmic}
\State {\bf input:} $C$, $\lambda$
\State Initialize set of randomized observations $\mathcal O = \varnothing$
\vspace{1em}
\For{$t=1,2,\dots$}
\If{$\abs{\mathcal O} \leq C r(m+n) \log(t)$}
\State \textit{Explore:}
  \State \hspace{\algorithmicindent} choose $S_t$ uniformly at random from all subsets of size $K$,
  \State \hspace{\algorithmicindent} observe customer type $i_t$,
  \State \hspace{\algorithmicindent} observe customer product choice $j_t$, and
  \State \hspace{\algorithmicindent} update the set of randomized observations $\mathcal O \leftarrow \mathcal O \cup (i_t, j_t, S_t)$.

\State \textit{Estimate:} let
\[
L(\Theta)=\frac1{|\mathcal O|}\sum_{(i,j,S) \in \mathcal O} \log\prns{\prns{1+\sum_{j'\in S}e^{\Theta_{ij'}}}\prns{\left\{\ba{lc}1 \quad\quad& j=0\\e^{\Theta_{ij}} \quad\quad& \text{otherwise}\ea\right.}^{-1}},
\]
\State \phantom{\textit{Estimate:}} and solve
\begin{equation*}
\ba{rl}
\widehat\Theta\in\operatornamewithlimits{argmin} \quad& L(\Theta) + \lambda\|\Theta\|_* \\
\text{s.t.} \quad& \|\Theta\|_\infty\leq\alpha/\sqrt{mn}
\ea
\end{equation*}
\Else
\State \textit{Exploit:}
choose $S_t\in S^\star(W,\widehat\Theta,\hat\mu;K)$.
\EndIf
\EndFor
\vspace{1em}
\State {\bf output:} $S_1, S_2, \ldots$
\end{algorithmic}
\end{algorithm}

Next we show that this achieves regret that is order-of-magnitude smaller than a structure-ignorant algorithm when structure is present.
Let
$$\delta(W,\Theta,\mu;K)=F(S^\star(W,\Theta,\mu;K);W,\Theta,\mu)-\max_{S\notin S^\star(W,\Theta,\mu;K) : \abs{S}\leq K}F(S;W,\Theta,\mu)$$
be the gap in revenue between the optimal assortment and any suboptimal assortment under parameter matrix $\Theta$.

\begin{theorem}\label{thm:regret-dyn}
Let
$r$ be such that $r\geq\op{rank}(\Theta^*)$,
$\rho\geq1$ be such that $1/\rho\leq m\mu_i\leq \rho$ $\forall i=1,\dots,m$,
$\alpha$ be such that $\magd{\Theta^\star}_\infty\leq\alpha/\sqrt{mn}$, $\delta$ be such that $\min_{1\leq i\leq m}\delta(W_i,\,\Theta^\star_i;\,K)\geq\delta$, and $\omega$ be such that $\magd{W}_\infty\leq\omega$.
Choose algorithm parameters as in eq.~\eqref{banditalgoparams}.
Then the regret of $\pi_\text{nuc-norm-plan}(C,\,\lambda)$
satisfies
\begin{align*}
\op{Regret}\prns{T;\pi_\text{nuc-norm-plan}(C,\,\lambda)}&\leq (C r(m+n)\log(T)+4)\omega\\&=O\prns{r\max(m,n)\log(T)}
\end{align*}
for all $T\leq (m+n)^{\frac{mn}{C(m+n)}r}$.
\end{theorem}\eeditb
%
\beditb\section{Proof of offline recovery guarantee}
We first introduce some notation to be used throughout our proofs.
Let $e_\ell$ be the $\ell\thh$ unit vector, and let $e_0$ is the vector of all zeros.
We infer the dimension of these vectors from context, so $e_{i_t}\in\R m$, $e_{j_t}\in\R n$.
Define
\begin{itemize}
\item the error to bound $\Delta=\widehat\Theta-\Theta^\star$,
\item the items not chosen $S_t'=S_t\backslash \{j_t\}$,
\item $\gamma=2\alpha/\sqrt{mn}$, a bound on the preference parameters and therefore on the error: $\|\Delta\|_\infty\leq \gamma$,
\item the selection indicator $X_{tj}=e_{i_t}e_j^T$, and
\item \editb{the mean square error on the $t$th observation $Y_t(\Delta) = \frac{1}{K_t}\sum_{j_t\in S_t}\Delta_{i_tj}^2$}
\end{itemize}
Using this notation, we can calculate the loss, its gradient, and its Hessian as
\begin{align*}
L(\Theta)&=\frac1N\sum_{t=1}^N\prns{\log\prns{1+\sum_{j\in S_t}e^{X_{tj}\cdot\Theta}}-X_{tj_t}\cdot\Theta}\\
\nabla L(\Theta)&=\frac1N\sum_{t=1}^N\prns{\frac{\sum_{j\in S_t}e^{X_{tj}\cdot\Theta}X_{tj}}{1+\sum_{j\in S_t}e^{X_{tj}\cdot\Theta}}-X_{tj_t}}\\
\nabla^2 L(\Theta)&=\frac{1}{N}\sum_{t=1}^N\frac{\prns{1+\sum_{j\in S_t}e^{X_{tj}\cdot\Theta}}\prns{\sum_{j\in S_t}e^{X_{tj}\cdot\Theta}X_{tj}^{\otimes2}}-\prns{\sum_{j\in S_t}e^{X_{tj}\cdot\Theta}X_{tj}}^{\otimes2}}{\prns{1+\sum_{j\in S_t}e^{X_{tj}\cdot\Theta}}^2},
\end{align*}
where $A^{\otimes2}=A\otimes A$ is the symmetric linear operator on matrices
defined by $(A\otimes A)(B)=(A\cdot B)A$.

\beditb
The proof of our main theorem makes use of a few lemmas, which we state here.
We defer the proofs of these lemmas until after the proof of the main theorem.

Our first step will be to show a quadratic lower bound $L_\textup{quad}(\Delta)$
on the Bregman divergence of our loss function around $\Theta^\star$,
\[
D_{\Theta^\star}(\Delta) = L(\Theta^\star + \Delta)-L(\Theta^\star)-\nabla L(\Theta^\star)\cdot \Delta
\]
Notice the Bregman divergence $D_{\Theta^\star}(\Delta)$ has the same Hessian as our loss function.
We will later prove (restricted) strong convexity of this lower bound,
which shows (restricted) strong convexity of our loss function.

\begin{lemma}[Quadratic lower bound]\label{lem:quad-lower}

Define the quadratic function
\[
L_\textup{quad}(\Delta) = \frac 1 N \sum_{t=1}^N Y_t(\Delta)
= \frac 1 N \sum_{t=1}^N \frac{1}{K_t}\sum_{j_t\in S_t}\Delta_{i_tj}^2.
\]
This function provides a lower bound on the Bregman divergence $D_{\Theta^\star}(\Delta)$
of our loss function around $\Theta^\star$:
\[
D_{\Theta^\star}(\Delta)
\geq \frac{1}{e^{4\gamma}}\frac{1}{4K}L_\textup{quad}(\Delta).
\]

\end{lemma}

We will then proceed to show that this quadratic lower bound $L_\textup{quad}(\Delta)$
is strongly convex, when restricted to relevant matrices $\Delta$, with high probability.
\eeditb

\begin{lemma}[Strong convexity]\label{lem:strongly-convex}
Fix a parameter $\tau\geq1$.
Let $$\mathcal A^\star=\braces{\Delta:~
\|\Delta\|_\infty\leq\gamma,\,
\|\Delta\|_*\leq\frac1{{\max\braces{(18\tau)^{1/4},\,480}}}\frac{1}{\sqrt{\rho^3Kmn}\gamma}\sqrt{\frac{N}{\sqrt{mn}\log (m+n)}}\|\Delta\|_\text{F}^2}.$$
We have
$$
\Prb{L_\textup{quad}(\Delta)\geq\frac1{2 \rho mn}\magd{\Delta}_\text{F}^2,
\quad\forall\Delta\in\mathcal A^\star}
\geq 1-2(m+n)^{-\tau}.
$$
\end{lemma}

\beditb
Lemma \ref{lem:strongly-convex} 
shows that $L_\textup{quad}(\Delta)$ is strongly convex with high probability
on a set of matrices $\mathcal A^\star$.
The argument $\Delta$ models the difference between
some parameter matrix $\Theta$ and the true parameter matrix $\Theta^\star$,
so it is reasonable to restrict our attention to error matrices $\Delta$ that
are in some sense small.
Here, the set of matrices $\mathcal A^\star$ is parametrized by
the maximum absolute value of any matrix entry, $\gamma$,
and a parameter $\tau$ which controls the size of the set.
The probability of strong convexity on this set is tunable using the parameter $\tau$,
and increases with $m$ and $n$.
\eeditb


\begin{lemma}[Near-optimality]\label{lem:truth-near-opt}
Fix a parameter $\tau\geq1$.
With probability at least $1-(m+n)^{-\tau}$,
$$
\magd{\nabla L(\Theta^\star)}_2\leq 4\sqrt{\tau}\sqrt{\frac{\rho K(m+n)\log (m+n)}{mnN}}.
$$
\end{lemma}

\beditb
The gradient of the loss $L(\Theta)$ vanishes for the maximum likelihood estimator $\Theta$.
Lemma \ref{lem:truth-near-opt} shows that, with high probability,
the true parameter matrix $\Theta^\star$
is nearly a maximum likelihood estimator for our problem,
since the gradient of $L(\Theta)$ nearly vanishes at $\Theta = \Theta^\star$.
The error we allow decreases with the problem dimensions $m$ and $n$, and
is controlled by a parameter $\tau$.
The probability that $\Theta^\star$ is nearly optimal
is again tunable using the parameter $\tau$,
and increases with $m$ and $n$.

To put these results together, we leverage the structure of our regularized objective and our choice of $\lambda$, as summarized by the following two lemmas.
\begin{lemma}[Regularized objective]\label{regularizedobjectivelemma}
We have that
\begin{equation}\label{Dupperbound}
D_{\Theta^\star}(\Delta)\leq
\magd{\nabla L(\Theta^\star)}_2\magd{\Delta}_*+\lambda\prns{\|{\Theta^\star}\|_*-\|{\widehat\Theta}\|_*}\leq\prns{\magd{\nabla L(\Theta^\star)}_2+\lambda}\magd{\Delta}_*.
\end{equation}
\end{lemma}
\begin{lemma}[Spectral decomposition]\label{spectraldecompositionlemma}
If $\magd{\nabla L(\Theta^\star)}\leq\lambda/2$, then, for any $r$,
\begin{equation}\label{eq-max-frob-or-rho}
\magd{\Delta}_*\leq16\max\braces{\sqrt{r}\magd{\Delta}_\text{F},\|{\overline\Theta^\star_r}\|_*}.
\end{equation}
\end{lemma}
The bound \eqref{eq-max-frob-or-rho} improves when
$\Theta^\star$ is low rank or approximately low rank.
The (approximate) low rank of $\Theta^\star$ enters our main result via this bound.

With these lemmas in hand, we can prove our main result, Theorem \ref{mainthm}.
We will prove these lemmas after the proof of Theorem \ref{mainthm}.

\proof{Proof of Theorem \ref{mainthm}}
Assume for now that $\Delta \in \mathcal A^\star$.
With probability at least $1-3(m+n)^{-\tau}$, the events in both
Lemma \ref{lem:strongly-convex} and Lemma \ref{lem:truth-near-opt} occur.

We next bound $D_{\Theta^\star}(\Delta)$ above and below.
By Lemma \ref{lem:truth-near-opt} and our choice of $\lambda$, we have $\magd{\nabla L(\Theta^\star)}_2\leq\lambda/2\leq\lambda$. Therefore, by Lemma~\ref{regularizedobjectivelemma}, $D_{\Theta^\star}(\Delta)\leq2\lambda\magd{\Delta}_*$.
%
Together with our choice of $\lambda$, this provides a upper bound on $D_{\Theta^\star}(\Delta)$.
A lower bound is given by Lemma \ref{lem:strongly-convex} and Lemma~\ref{lem:quad-lower}.
Together, this yields
\begin{equation}
\label{eq-rsc}
\frac1{8e^{4\gamma}\rho mnK}\magd{\Delta}_\text{F}^2
\leq D_{\Theta^\star}(\Delta) \leq
16\sqrt{\tau}
\sqrt{\frac{\rho K(m+n)\log (m+n)}{mnN}}
\magd{\Delta}_*.
\end{equation}
Hence recalling $\gamma = \frac{2\alpha}{\sqrt{mn}}$,
\begin{equation}
\label{eq-fro-lt-nuc}
\magd{\Delta}_\text{F}^2\leq 256\sqrt{\tau}\alpha e^{\frac{8\alpha}{\sqrt{mn}}} (\rho K)^{3/2}\sqrt{\frac{(m+n)\log (m+n)}{N}}\magd{\Delta}_*.
\end{equation}
\beditb
The bound \eqref{eq-fro-lt-nuc}, which shows that the square Frobenius norm of the error
is controlled by its nuclear norm,
establishes the most important part of our proof.

We next show that \eqref{eq-fro-lt-nuc} holds even if $\Delta \not \in \mathcal A^\star$.
Suppose so, \ie,
\[
\|\Delta\|_*>\max\braces{(18\tau)^{1/4},\,480}\frac{1}{\sqrt{\rho^{3} Kmn}\gamma}\sqrt{\frac{N}{\sqrt{mn}\log (m+n)}}\|\Delta\|_\text{F}^2.
\]
Rewriting and introducing redundant terms greater than 1,
we recover \eqref{eq-fro-lt-nuc}. This establishes \eqref{eq-fro-lt-nuc} holds for all $\Delta$, with high probability.

To show our main result, we will bound the nuclear norm of the error in terms of the Frobenius norm (not squared).
Dividing both sides of \eqref{eq-fro-lt-nuc} by the Frobenus norm of the error will yield the result.
This is achieved by leveraging Lemma~\ref{spectraldecompositionlemma}.
If $\sqrt{r}\|\Delta\|_\text{F}\geq \|{\overline\Theta^\star_r}\|_*$,
substitute \eqref{eq-max-frob-or-rho} in \eqref{eq-fro-lt-nuc}
to see
$$\magd{\Delta}_\text{F}\leq 4096\sqrt{\tau}\alpha e^{\frac{8\alpha}{\sqrt{mn}}} K^{3/2}\sqrt{\frac{r(m+n)\log (m+n)}{N}}.$$
Otherwise (if $\sqrt{r}\|\Delta\|_\text{F}< \|{\overline\Theta^\star_r}\|_*$),
substitute \eqref{eq-max-frob-or-rho} in \eqref{eq-fro-lt-nuc}
and take the square root to see
\begin{align*}
\magd{\Delta}_\text{F}&\leq\sqrt{4096\sqrt{\tau}\alpha e^{\frac{8\alpha}{\sqrt{mn}}} K^{3/2}\sqrt{\frac{\|{\overline\Theta^\star_r}\|_* (m+n)\log (m+n)}{N}}}\\&\leq 4096\sqrt{\tau}\alpha e^{\frac{8\alpha}{\sqrt{mn}}} K^{3/4}\prns{{\frac{\|{\overline\Theta^\star_r}\|_* d\log d}{N}}}^{1/4}.
\end{align*}
Combining yields the statement.\eeditb
\halmos\endproof

\proof{Proof of Lemma~\ref{lem:quad-lower}}
Define $\Delta = \Theta - \Theta^\star$.
By Taylor's theorem, there is some $s\in[0,1]$ such that
\begin{align*}
L(\widehat \Theta)-L(\Theta^\star)-\nabla L(\Theta^\star)\cdot\Delta
&=\nabla^2L(\Theta^\star+s\Delta)[\Delta,\Delta]
\\&=\frac1N\sum_{t=1}^N\frac{\prns{1+\sum_{j\in S_t}e^{v_{tj}}}\prns{\sum_{j\in S_t}e^{v_{tj}}(X_{tj}\cdot\Delta)^2}-\prns{\sum_{j\in S_t}e^{v_{tj}}X_{tj}\cdot\Delta}^2}{\prns{1+\sum_{j\in S_t}e^{v_{tj}}}^2}
\\&\geq\frac1N\sum_{t=1}^N\frac{\sum_{j\in S_t}e^{v_{tj}}(X_{tj}\cdot\Delta)^2}{\prns{1+\sum_{j\in S_t}e^{v_{tj}}}^2}\geq\frac1N\frac{1}{e^{4\gamma}}\sum_{t=1}^N\frac{1}{(K_t+1)^2}\sum_{j\in S_t}(X_{tj}\cdot\Delta)^2
\\&\geq\frac1N\frac{1}{e^{4\gamma}}\frac{1}{4K}\sum_{t=1}^NY_t(\Delta),
\\&=\frac{1}{e^{4\gamma}}\frac{1}{4K} L_\textup{quad}(\Delta),
%
%
\end{align*}
where $v_{tj}=X_{tj}\cdot(\Theta^\star+s\Delta)$, the first inequality is Jensen's, and the second is from $\abs{v_{tj}}\leq 2\gamma$.
\halmos\endproof

\beditb
Lemma \ref{lem:strongly-convex} builds on Lemma \ref{lem:strongly-convex-bounded},
which we now state and prove.

\begin{lemma}\label{lem:strongly-convex-bounded}
Let $\mathcal A_{\Gamma,\nu}=
\{\Delta:~
\|\Delta\|_\infty\leq\gamma,\,
\|\Delta\|_\text{F}\leq\Gamma,\,
\|\Delta\|_*\leq\frac{\nu}{60 \sqrt{\rho^3Kmn}\gamma}\sqrt{\frac{N}{(m+n)\log (m+n)}}\Gamma^2\}$.
Define the maximum deviation from strong convexity
$$\mathcal M_{\Gamma,\nu}=\sup_{\Delta\in\mathcal A_{\Gamma,\nu}}
\prns{\frac1{\rho mn}\|\Delta\|_\text{F}^2 - L_\textup{quad}(\Delta)}.$$
Then
$$\Prb{\mathcal M_{\Gamma,\nu}\geq \nu\frac{\Gamma^2}{\rho mn}}
\leq\exp\prns{-{\frac{8}{9}}\frac{\nu^2}{\rho^2m^2n^2}\frac{\Gamma^4}{\gamma^4}N}.$$
\end{lemma}

\beditb
Lemma \ref{lem:strongly-convex-bounded} provides our first steps towards
showing that our log likelihood objective $L(\Theta)$ is strongly convex, in a restricted sense.
It shows that,
on a certain bounded set of matrices $\mathcal A_{\Gamma,\nu}$,
the quadratic lower bound $L_\textup{quad}(\Delta)$ is
unlikely to be very far from strongly convex.
The bound is parametrized by $\gamma$, which bounds the infinity norm,
and $\Gamma$, which bounds the Frobenius and nuclear norms of the matrix.
We see that as the problem gets larger ($m$, $n$ and $N$ increase),
the distance to strong convexity ($\nu\frac{\Gamma^2}{\rho mn}$) decreases,
and the probability the bound fails ($\exp\prns{-{\frac{8}{9}}\frac{\nu^2}{\rho^2m^2n^2}\frac{\Gamma^4}{\gamma^4}N}$) decreases.
\eeditb

\proof{Proof of Lemma \ref{lem:strongly-convex-bounded}}

Since $S_t$ is symmetric, we have that $$\Eb{Y_t(\Delta)}=\sum_{i=1}^m\mu_i\sum_{j=1}^n\frac1n\Delta_{ij}^2\geq\frac{1}{\rho mn}\magd{\Delta}_F^2.$$
Define
$$
\tilde{\mathcal M}_{\Gamma,\nu}
={\sup_{\Delta\in\mathcal A_{\Gamma,\nu}}{\frac1N\sum_{t=1}^N\prns{\E Y_t(\Delta)-Y_t(\Delta)}}}
$$
and note that $\tilde{\mathcal M}_{\Gamma,\nu}\geq {\mathcal M}_{\Gamma,\nu}$.
Then, letting $Y'_t(\Delta)$ be an identical and independent replicate of $Y_t(\Delta)$ and letting $\epsilon_t$ be iid Rademacher random variables independent of all else, we have
\begin{align*}
\E{\tilde{\mathcal M}}_{\Gamma,\nu}
&=\Eb{\sup_{\Delta\in\mathcal A_{\Gamma,\nu}}{\frac1N\sum_{t=1}^N\prns{\E Y'_t(\Delta)-Y_t(\Delta)}}}
\leq\Eb{\sup_{\Delta\in\mathcal A_{\Gamma,\nu}}{\frac1N\sum_{t=1}^N\prns{Y'_t(\Delta)-Y_t(\Delta)}}}\\
&=\Eb{\sup_{\Delta\in\mathcal A_{\Gamma,\nu}}{\frac1N\sum_{t=1}^N\epsilon_t\prns{Y'_t(\Delta)-Y_t(\Delta)}}}
\leq2\Eb{\sup_{\Delta\in\mathcal A_{\Gamma,\nu}}{\frac1N\sum_{t=1}^N\epsilon_tY_t(\Delta)}}\\
&=2\Eb{\sup_{\Delta\in\mathcal A_{\Gamma,\nu}}{\frac1N\sum_{t=1}^N\epsilon_t\frac{1}{K_t}\magd{e_{i_t}^T\Delta \sum_{j\in S_t}e_j}_2^2}}.
\end{align*}

Note that
$$\sup_{v,v'\in[-\gamma,\gamma]^k\times\{0\}^{n-k}}\frac{\magd v_2^2-\magd{v'}_2^2}{\magd{v-v'}_\infty}=2\gamma k,$$
i.e., $\magd\cdot_2^2$ is $2\gamma k$-Lipschitz with respect to the $\infty$-norm on a domain in $[-\gamma,\gamma]^n$ where only $k$ entries are nonzero.
Therefore, by Lemma 7 of \cite{bertsimas2014predictive} and by H\"older's inequality, letting $W_t=\sum_{j\in S_t}\epsilon_{tj}e_{i_t}e_{j}^T$ where $\epsilon_{tj}$ are new iid Rademacher random variables independent of all else,
\begin{align*}
\E{\tilde{\mathcal M}}_{\Gamma,\nu}
&\leq4\gamma
\Eb{
\sup_{\Delta\in\mathcal A_{\Gamma,\nu}}
\frac1N
\sum_{t=1}^N
W_t\cdot\Delta
}
\leq 4\gamma
\E\magd{\frac1N\sum_{t=1}^NW_t}_2
\sup_{\Delta\in\mathcal A_{\Gamma,\nu}}\magd{\Delta}_*.
\end{align*}
Note that
\begin{align*}
\magd{W_t}_2&\leq\sqrt K\\
\Eb{W_tW_t^T\big|S_t,i_t}&=K_te_{i_t}e_{i_t}^T&&\text{ and hence }&&\magd{\Eb{W_tW_t^T}}_2\leq \rho K/m\\
\Eb{W_t^TW_t\big|S_t,j_t}&=\sum_{j\in S_t}e_{j}e_{j}^T&&\text{ and hence }&&\magd{\Eb{W_t^TW_t}}_2\leq K/n
\end{align*}

Hence, the matrix Bernstein inequality (Theorem 1.6 of \cite{tropp2012user}) gives that $\magd{\frac1N\sum_{t=1}^NW_t}_2\geq \delta$ with probability at most
$$
(m+n)\max\braces{e^{-\frac{N\delta^2\min\braces{m,n}}{4\rho K}},\,e^{-\frac{\delta}{\sqrt K}}}.
$$
Setting the probability to $1/\sqrt{mn\min\{m,n\}}$ and using $N\leq mn\log (m+n)$,
\begin{align*}
\E\bracks{\magd{\frac1N\sum_{t=1}^NW_t}_2}
&\leq \sqrt{\frac{K}{mn\min\{m,n\}}}+2\sqrt{\frac{\rho K\prns{\frac12\log(m)+\frac12\log(n)\frac12\log(\min\{m,n\})+\log(m+n)}}{N\min\{m,n\}}}\\
&\leq
\sqrt{\frac{K\log(m+n)}{N\min\{m,n\}}}+2\sqrt{\frac{\frac52\rho K\log(m+n)}{N\min\{m,n\}}}
\leq 5\sqrt{\frac{\rho K\log(m+n)}{N\min\{m,n\}}}
\end{align*}
Putting it all together, we get,
$$
\E{\tilde{\mathcal M}}_{\Gamma,\nu}\leq 20\gamma \sqrt{\frac{\rho K\log(m+n)}{N\min\{m,n\}}}\frac{\nu}{60 \sqrt{\rho^3Kmn}\gamma}\sqrt{\frac{N}{(m+n)\log (m+n)}}\Gamma^2\leq\frac{\nu}{3}\frac{\Gamma^2}{\rho mn}.
$$

Next we use this to prove the concentration of $\tilde{\mathcal M}_{\Gamma,\nu}$.
Let $\tilde{\mathcal M}'_{\Gamma,\nu}$ be a replicate of $\tilde{\mathcal M}_{\Gamma,\nu}$ with $i'_t=i_t,\,S'_t=S_t$ for all $t$ except $t'$. Then the difference $\tilde{\mathcal M}_{\Gamma,\nu}-\tilde{\mathcal M}'_{\Gamma,\nu}$ is bounded by
\begin{align*}
&\frac1N\sup_{\Delta\in\mathcal A_{\Gamma,\nu}}\prns{
\frac{1}{K_{t'}}\sum_{j\in S_{t'}}\Delta_{i_{t'}j}^2-\frac{1}{K_{t'}}\sum_{j\in S'_{t'}}\Delta_{i'_{t'}j}^2
}\leq \frac1N\prns{\gamma^2-0}= \frac{\gamma^2}N.
\end{align*}
Hence, by McDiarmid's inequality, we have
$$\Prb{\tilde{\mathcal M}_{\Gamma,\nu}\geq \nu\frac{\Gamma^2}{\rho mn}}
\leq\Prb{\tilde{\mathcal M}_{\Gamma,\nu}-\E{\mathcal M}_{\Gamma,\nu}\geq \frac{2\nu}{3}\frac{\Gamma^2}{\rho mn}}\leq \exp\prns{-\frac{8}{9}\frac{N\nu^2\Gamma^4}{\gamma^4\rho^2m^2n^2}},$$
which yields the result, given $\tilde{\mathcal M}\geq\mathcal M$.
\halmos\endproof

\proof{Proof of Lemma \ref{lem:strongly-convex}}

\beditb
Lemma \ref{lem:strongly-convex} extends Lemma \ref{lem:strongly-convex-bounded}
to show that $L_\textup{quad}(\Delta)$ is strongly convex with high probability
on a larger set of matrices $\mathcal A^\star$.
Compared to Lemma \ref{lem:strongly-convex-bounded}, we eliminate the nuisance parameter $\Gamma$,
using a peeling argument as in \citet{raskutti2010restricted}.
\eeditb

Let $\tau'=K^2\rho^4\max\braces{\tau,\,480^4/18}$.
Since $\|\cdot\|_*\geq\|\cdot\|_\text{F}$, we have $\inf_{\Delta\in\mathcal A^\star:\Delta\neq0}\|\Delta\|_\text{F}\geq\eta:=\prns{18\tau'}^{1/4}\sqrt{\rho mn}\gamma\sqrt{\sqrt{mn}\log (m+n)/N}$.
Set $\nu=60(18\tau')^{-1/4}$ and $\beta=\sqrt{1/(2\nu)}$.
Since $\tau'\geq480^4/18$ we have $\nu\leq1/8$ and $\beta\geq2>1$.
Let $\mathcal A_l=\mathcal A^\star\cap\{\eta\beta^{l-1}\leq\magd\Delta_\text{F}\leq\eta\beta^l\}$ and note that $\mathcal A^\star=\bigcup_{l=1}^\infty A_l$ and $\mathcal A_l\subset \mathcal A_{\beta^l\eta,\nu}$.

If $\Delta\in \mathcal A_l$ has $L_\textup{quad}(\Delta)<\frac1{2 \rho mn}\magd{\Delta}_\text{F}^2$ then
\begin{align*}
&\frac1{\rho mn}\magd{\Delta}_\text{F}^2-L_\textup{quad}(\Delta)>\frac1{2\rho mn}\magd{\Delta}_\text{F}^2
\geq\frac1{2 \rho mn}\prns{\beta^{l-1}\eta}^{2}=\frac1{ \rho mn}\frac{1}{2\beta^2}(\beta^l\eta)^2=\frac\nu{\rho mn}(\beta^l\eta)^2.
\end{align*}
Therefore, the probability that the event in the statement of theorem is invalid is bounded by
\begin{align*}
\min\braces{1, \sum_{l=1}^\infty\Prb{\mathcal M_{\beta^l\eta,\nu}>\frac\nu{\rho mn}\prns{{\beta^\ell\eta}}^2}}
&\leq\min\braces{1, \sum_{l=1}^\infty\exp\prns{-\frac{8}{9}\frac{\nu^2\beta^{4\ell}\eta^4N}{\rho^2m^2n^2\gamma^4}}}\\
&\leq\min\braces{1, \sum_{l=1}^\infty\exp\prns{-\frac{1}{72}\frac{4^{\ell}\eta^4N}{\rho^2m^2n^2\gamma^4}}}\\
&\leq \min\braces{1,\sum_{l=1}^\infty\exp\prns{-\frac1{72} \frac{4^{l}\eta^4 N}{\rho^2m^2n^2\gamma^4}}}\\
&\leq \min\braces{1,\sum_{l=1}^\infty\exp\prns{-\frac1{18} \frac{\eta^4 N}{\rho^2m^2n^2\gamma^4}l}}\\
&=\min\braces{1,\prns{\exp\prns{\frac1{18} \frac{\eta^4 N}{\rho^2m^2n^2\gamma^4}}-1}^{-1}}\\
&\leq2\exp\prns{-\frac1{18} \frac{\eta^4 N}{\rho^2m^2n^2\gamma^4}}\\
&=2\exp\prns{-\tau' mn(\log (m+n))^2/N}\\
&\leq 2 (m+n)^{-\tau'}\leq 2 (m+n)^{-\tau},
\end{align*}
using Lemma \ref{lem:strongly-convex-bounded} and $N\leq mn\log (m+n)$.
\halmos\endproof

\proof{Proof of Lemma \ref{lem:truth-near-opt}}
Let
$\psi_{tj}=e^{X_{tj}\cdot\Theta^\star}$ and
$G_t=\frac{\sum_{j\in S_t}\psi_{tj}X_{tj}}{1+\sum_{j\in S_t}\psi_{tj}}-X_{tj_t}$ so that $\nabla L(\Theta^\star)=\frac1N\sum_{t=1}^NG_t$.
Note that because $j_t$ is drawn according to $\Theta^\star$ and $X_{t0}=0$, we have that $\Eb{G_t\big|i_t,S_t}=0$ and hence $\E G_t=0$.
Note that $\|G_t\|_2\leq\sqrt{2}$. Moreover,
$$
G_tG_t^T=e_{i_t}e_{i_t}^T\prns{ 1 - \frac{2\psi_{tj_t}}{1+\sum_{j\in S_t}\psi_{tj}} + \frac{\sum_{j\in S_t}\psi_{tj}^2}{\prns{1+\sum_{j\in S_t}\psi_{tj}}^2} }.
$$
Since by Jensen's inequality the multiplier in the parentheses is no greater than 2, we get $\magd{\Eb{G_tG_t^T}}_2\leq\frac{2\rho}m\leq\frac{2\rho K}{m}$. Letting $y_{tj}=\indic{j=j_t}$, we have
$$
G_t^TG_t=\hspace{-1em}\sum_{j\in S_t\\k\in S_t} e_je_k^T\prns{ y_{tj}y_{tk} - \frac{2y_{tj}\psi_{tk}}{1+\sum_{l\in S_t}\psi_{tl}} + \frac{\psi_{tj}\psi_{tk}}{\prns{1+\sum_{l\in S_t}\psi_{tl}}^2} }.
$$
Noting that $y_{tj}\geq0$, $\psi_{tj}\geq0$, and $y_{tj}y_{tk}\leq\indic{j=k}$, we see that
\begin{align*}
\magd{\Eb{G_t^TG_t}}_2&\leq \magd{\Eb{\sum_{j\in S_t}e_je_j^T}}_2 + \magd{{\Eb{\frac1{{K_t^2}}\prns{\sum_{j\in S_t}e_j}\prns{\sum_{j\in S_t}e_j}^T}}}_2
\leq \frac Kn+\frac1n\leq \frac{2\rho K}n.
\end{align*}
Hence, by the matrix Bernstein inequality (Theorem 1.6 of \cite{tropp2012user}),
$$
\magd{\frac1N\sum_{t=1}^NG_t}_2\leq2\sqrt{2(\tau+1)}\sqrt{\frac{\rho K\log (m+n)}{\min\braces{m,n}N}}\leq4\sqrt{\tau}\sqrt{\frac{\rho K(m+n)\log (m+n)}{mnN}},
$$
with probability at least $1-(m+n)^{-\tau}$.
\halmos\endproof
\beditb
\proof{Proof of Lemma \ref{regularizedobjectivelemma}}
By the optimality of $\widehat\Theta$, we have
$$
L(\widehat\Theta)+\lambda\|{\widehat\Theta}\|_*\leq L(\Theta^\star)+\lambda\|{\Theta^\star}\|_*.
$$
By H\"older's inequality,
\begin{align}
D_{\Theta^\star}(\Delta)&=L(\widehat\Theta)-L(\Theta^\star)-\nabla L(\Theta^\star)\cdot\Delta\notag\\
&\leq
\magd{\nabla L(\Theta^\star)}_2\magd{\Delta}_*+\lambda\prns{\|{\Theta^\star}\|_*-\|{\widehat\Theta}\|_*},\notag
\end{align}
yielding the first inequality in the statement. The second inequality in the statement follows by triangle inequality.
\halmos\endproof

\proof{Proof of Lemma \ref{spectraldecompositionlemma}}
We introduce a simple linear algebraic decomposition of $\Delta$
in terms of the principal subspaces of $\Theta^*$,
following the method of \cite{recht2010}.
Let $\Theta^\star=U\op{Diag}(\sigma_1,\sigma_2,\dots)V^T$ be the singular-value decomposition
of $\Theta^\star$ with singular values sorted largest to smallest.
Using block notation, define $\Gamma$ by
\[
U^T\Delta V=\Gamma=\prns{\begin{array}{cc}\Gamma_{11}&\Gamma_{12}\\\Gamma_{21}&\Gamma_{22}\end{array}}\ \text{with $\Gamma_{11}\in\R{r\times r}$}
\]
and define $\Delta'$ and $\Delta''$ as
\[
\Delta''=U\prns{\begin{array}{cc}0&0\\0&\Gamma_{22}\end{array}}V^T,
\qquad
\Delta'=U\prns{\begin{array}{cc}\Gamma_{11}&\Gamma_{12}\\\Gamma_{21}&0\end{array}}V^T,
\]
so $\Delta=U\Gamma V^T=\Delta'+\Delta''$.
We bound the rank of $\Delta'$ as
\begin{align*}\op{rank}&(\Delta')=\op{rank}(U^T\Delta'V)
\\=&~\op{rank}\prns{
\prns{\begin{array}{cc}\Gamma_{11}/2&\Gamma_{12}\\0&0\end{array}}
+
\prns{\begin{array}{cc}\Gamma_{11}/2&0\\\Gamma_{21}&0\end{array}}
}\leq 2r.
\end{align*}
Define the restriction of $\Theta$ to its principal subspace,
$\Theta^\star_r=U\op{Diag}(\sigma_1,\dots,\sigma_{r},0,0,\dots)V^T$,
and its complement $\overline\Theta^\star_r=\Theta^\star-\Theta^\star_r$.
Bound the nuclear norm of $\widehat \Theta$ as
\begin{align*}
\|{\widehat\Theta}\|_*&=\|{\Theta^\star+\Delta}\|_*=\|{\Theta^\star_r+\overline\Theta^\star_r+\Delta'+\Delta''}\|_*\\
&\geq \|{\Theta^\star_r+\Delta''}\|_*-\|{\overline\Theta^\star_r}\|_*-\|{\Delta'}\|_*\\
&=\|{\Theta^\star_r}\|_*+\|{\Delta''}\|_*-\|{\overline\Theta^\star_r}\|_*-\|{\Delta'}\|_*\\
&=\|{\Theta^\star}\|_*+\|{\Delta''}\|_*-2\|{\overline\Theta^\star_r}\|_*-\|{\Delta'}\|_*,
\end{align*}
and so we arrive at the inequality
\begin{equation}
\label{eq:linear-algebra}
\|{\widehat\Theta}\|_*-\|{\Theta^\star}\|_*\leq 2\|{\overline\Theta^\star_r}\|_*+\|{\Delta'}\|_*-\|{\Delta''}\|_*.
\end{equation}
Since $\magd{\nabla L(\Theta^\star)}_2\leq\lambda/2$, by assumption, eq.~\eqref{Dupperbound} and the nonnegativity of the Bregman divergence yield
\begin{align*}
0&\leq
\|{\nabla L(\Theta^\star)}\|_2\|{\Delta}\|_*+\lambda\prns{\|{\Theta^\star}\|_*-\|{\widehat\Theta}\|_*}\\
&\leq\lambda\prns{\frac12\|{\Delta}\|_*+2\|{\overline\Theta^\star_r}\|_*+\|{\Delta'}\|_*-\|{\Delta''}\|_*}\\
&\leq\lambda\prns{2\|{\overline\Theta^\star_r}\|_*+\frac32\|{\Delta'}\|_*-\frac12\|{\Delta''}\|_*},
\end{align*}
using \eqref{eq:linear-algebra} for the first inequality.

Reorganizing, we see
\begin{equation}\label{eq:error-subspace}
\magd{\Delta''}_*\leq 3\magd{\Delta'}_*+4\|{\overline\Theta^\star_r}\|_*.
\end{equation}
This interesting inequality states that the error in our estimate
off of the principal subspace of $\Theta^\star$ is controlled by
(the sum of) the error \emph{on} the principal subspace and
the energy of $\Theta^\star$ outside of the principal subspace,
\[
\|{\overline\Theta^\star_r}\|_*=\sum_{j=r+1}^{\min\{m,n\}}\sigma_{j}.
\]
To arrive at this bound, we have used only
the optimality of $\widehat \Theta$ for our objective,
together with some basic linear algebra.

We can use \eqref{eq:error-subspace} to control the nuclear norm of $\Delta$:
\begin{align*}
\magd{\Delta}_*&\leq\magd{\Delta'}_*+\magd{\Delta''}_*=4\magd{\Delta'}_*+4\|{\overline\Theta^\star_r}\|_*
\\
&\leq8\max\braces{\magd{\Delta'}_*,\|{\overline\Theta^\star_r}\|_*}\\
&\leq8\max\braces{\sqrt{2r}\magd{\Delta'}_\text{F},\|{\overline\Theta^\star_r}\|_*}\\
&\leq8\max\braces{\sqrt{2r}\magd{\Delta}_\text{F},\|{\overline\Theta^\star_r}\|_*}\\
&\leq16\max\braces{\sqrt{r}\magd{\Delta}_\text{F},\|{\overline\Theta^\star_r}\|_*},
\end{align*}
where the second-to-last inequality uses the fact that $\magd{\Delta}_\text{F}-\magd{\Delta'}_\text{F}=\magd{\Gamma_{22}}_\text{F}\geq0$. This yields the second result.
\halmos\endproof

This lemma concludes the proof of Theorem \ref{mainthm}.
We now prove that the estimate $\hat \mu$ for the customer type
distribution is consistent.
\eeditb

\proof{Proof of Theorem \ref{muthm}} 
We prove this by cases, depending on which term in the min is smaller. Note that this is not a random event so we can choose which bound to use a priori, yielding the min.

First we deal with the second term. By union bound and Hoeffding's inequality,
$$
\Prb{\magd{\hat\mu-\mu^\star}_q>\eta}\leq \sum_{i=1}^m\Prb{\abs{\hat\mu_i-\mu_i^\star}>\eta/m^{1/q}}\leq 2m\exp(-2n\eta^2/m^{2/q})
$$

Next we deal with the first term.
Note that $\magd\cdot_q\leq\magd\cdot_1$ so it is sufficient to prove this for $q=1$.
Let $I_t=e_{i_t}$ be the type indicator observation
and let $I$ be a generic draw of $I_t$ from $\mu$. Then $\hat\mu=\frac1{N}\sum_{t=1}^NI_t$. By \cite{bartlett2003rademacher}, we have that with probability at least $1-\nu$,
$$
\magd{\mu-\hat\mu}_1=\sup_{\magd{v}_\infty\leq1}\prns{\Eb{v^TI}-\frac1{N}\sum_{t=1}^Nv^TI_t}\leq 2\widehat{\mathfrak R}_{N}+\sqrt{\frac{-\log\nu}{2N}},
$$
where $\widehat{\mathfrak R}_{N}$ is the empirical Rademacher complexity
$$
\widehat{\mathfrak R}_{N}=\frac{1}{2^{N}}\sum_{\epsilon\in\{-1,+1\}^{N}}\sup_{\magd v_\infty\leq1}\frac1{N}\sum_{t=1}^{N}\epsilon_tv^TI_t.
$$
By linearity and
duality of norms, we have
\begin{align*}
\widehat{\mathfrak R}_{N}&= \frac{1}{2^{N}}\sum_{\epsilon\in\{-1,+1\}^{N}}\sup_{\magd v_\infty\leq1}v^T\prns{\frac1{N}\sum_{t=1}^{N}\epsilon_tI_t}\\
&=
\frac{1}{2^{N}}\sum_{\epsilon\in\{-1,+1\}^{N}}\magd{\frac1{N}\sum_{t=1}^{N}\epsilon_tI_t}_1\\
&=\frac1{N}\sum_{i=1}^m\frac{1}{2^{N}}\sum_{\epsilon\in\{-1,+1\}^{N}}\abs{\sum_{t=1}^{N}\indic{i_t=i}\epsilon_t}\\
&=\frac{1}{N}\sum_{i=1}^m\frac{1}{2^{N\hat\mu_i}}\sum_{\epsilon\in\{-1,+1\}^{N\hat\mu_i}}\abs{\sum_{t=1}^{N\hat\mu_i}\epsilon_t}\\
&=\frac{1}{N}\sum_{i=1}^m\frac{1}{2^{N\hat\mu_i-1}}\left\lceil\frac {N\hat\mu_i}2\right\rceil\binom{m}{\left\lceil\frac {N\hat\mu_i}2\right\rceil}\\
&\leq\frac{1}{N}\sum_{i=1}^m\sqrt{N\hat\mu_i}=\frac{1}{\sqrt{N}}\sum_{i=1}^m\sqrt{\hat\mu_i}\leq\sqrt{\frac{{m}}{N}}.
\end{align*}
Therefore, using the concavity of square root, we have that
$$
\Prb{\magd{\mu-\hat\mu}_1>\eta}\leq e^{m-N\eta^2/64}.
$$
Rearranging yields the result.
\halmos\endproof

\subsection{Proofs Omitted from Section \ref{fgdalgorithm}}

\proof{Proof of Lemma \ref{equivprobs}}
Given $\Theta$ feasible in \eqref{eq-alg-alt} with $\op{rank}(\Theta)\leq\tilde r$, write its SVD
$\Theta=\tilde U\Sigma \tilde V^T$, where $\Sigma\in\R{\tilde r\times\tilde r}$ is diagonal and $U,\,V$ unitary. 
Letting $U=\tilde U\Sigma^{1/2}$ and $V=\tilde V\Sigma^{1/2}$,
we obtain a feasible solution to \eqref{eq-alg-alt2} with the same objective value
$\Theta$ has in \eqref{eq-alg-alt}.
Conversely, given $U,\,V$ feasible in \eqref{eq-alg-alt2}, let $\Theta=UV^T$.
Note $\op{rank}(\Theta)\leq\tilde r$ and
\begin{align*}
&\|\Theta\|_*\leq\|UV^T\|_*=\op{tr}(\Sigma)=\op{tr}(\tilde U^TUV^T\tilde V)\\&\leq\|\tilde U^TU\|_\text{F}\|V^T\tilde V\|_\text{F}\leq\|U\|_\text{F}\|V\|_\text{F}\leq\frac12\|U\|^2_\text{F}+\frac12\|V\|^2_\text{F}.
\end{align*}
Hence, $\Theta$ has objective value no worse than $(U, V)$.
\halmos\endproof

\subsection{Proofs of Regret Bounds}

Before presenting the proofs that a structure aware algorithm can achieve regret
sublinear in the number of matrix entries $mn$, we formally prove that
a structure ignorant algorithm must have regret linear in $mn$.
This result, Theorem~\ref{regcor}, makes use of
Theorem~1 in \cite{saure2013}, which we restate here in our notation:

\begin{theorem}[\cite{saure2013}, Theorem 1]\label{saure2013}
Consider a \edit{LRCMNL} with one type, $m=1$.
Under any consistent policy $\pi$ for choosing the sets $S_t$, 
\[
\E \op{Regret}_T(\pi) \geq C_1 + C_2 \tilde{\cal N} \log T 
\]
for all $T$,
where
\[
\tilde{\cal N} = \abs{\braces{j\notin S^\star(W_i,\Theta^\star_i;K):~
\exists\theta\in\R n,\,j\in S^\star(W_i,\theta;K),\,\theta_{j'}=\Theta^\star_{ij}~
\forall j'\in S^\star(W_i,\Theta^\star_i;K)}}
\]
is the set of potentially optimal items,
and $C_1$ and $C_2$ are constants independent of $n$ and $T$.
\end{theorem}



Note that \cite{saure2013} prove this lower bound only for
policies that do not depend on the time horizon $T$,
since their proof requires taking $T \to \infty$.
However, this proof is easily extended to a lower bound
even for policies that depend on the time horizon by
considering a sequence of policies for each $T$.

\proof{Proof of Theorem \ref{regcor}}

First, use a Chernoff bound to see that the number of times $T_i$ that
type $i$ is chosen is larger than $\frac{T \mu_i}{2}$ with high probability:
\[
\Prb{T_i > \frac{T \mu_i}{2}} \geq 1 - \exp\left(\frac{-1}{8}T \mu_i\right).
\]
Recall that $T = \Omega(m^{1+\epsilon})$ grows superlinearly in the number of types $m$. We use a union bound to show that for large $T$, with high probability
we have $T_i \geq \frac{T \mu_i}{2}$ for every $i$:
\begin{eqnarray*}
\Prb{T_i \geq \frac{T \mu_i}{2} \forall i}
&\geq& 1 - \sum_{i=1}^m \exp\left(\frac{-1}{8}T \mu_i\right) \\
&\geq& 1 - \sum_{i=1}^m \exp\left(\frac{-T}{8 \rho m}\right) \\
&\geq& 1 - \sum_{i=1}^m \exp\left(\frac{-m^\epsilon}{8\rho}\right) \\
&\geq& 1 - m\exp\left(\frac{-m^\epsilon}{8\rho}\right),
\end{eqnarray*}
which converges to 1 as $m$ increases.
In particular, for sufficiently large $T$ and $m$,
$\Prb{T_i \geq \frac{T \mu_i}{2} \forall i} > \frac 1 2$.
Let's suppose below that $T$ and $m$ are large enough that this bound holds.
\newcommand{\oftenrows}{T_i \geq \frac{T \mu_i}{2}~ \forall i}

Let $r(i;S)=\frac{\sum_{j\in S}\exp{\Theta^\star_{ij}}W_{ij}}{1+\sum_{j\in S}\exp{\Theta^\star_{ij}}}$ and let $R_t=\max_{\abs{S}\leq K}r(i_t;S)-r(i_t;S_t)$.
Then conditioning on the event that $T_i \geq \frac{T \mu_i}{2}$ for every $i$,
\begin{eqnarray*}
\E \op{Regret}_T(\pi)=\E\left[\sum_{t=1}^T R_t\right]
&\geq& \E\left[\sum_{t=1}^T R_t \mid \oftenrows \right]\Prb{\oftenrows} \\
&\geq& \frac12 \E\left[\sum_{t=1}^T R_t \mid \oftenrows \right] \\
&=& \frac12 \E\left[\sum_{i=1}^m \sum_{t:~ i_t = i} R_t \mid \oftenrows \right]\\
&=& \frac12 \left(\sum_{i=1}^m \E\left[\sum_{t:~ i_t = i} R_t \mid \oftenrows \right]\right) \\
&=& \frac12 \left(\sum_{i=1}^m \E\left[\sum_{t:~ i_t = i} R_t \mid T_i \geq \frac{T \mu_i}{2} \right]\right)
\end{eqnarray*}
Now, using Theorem~\ref{saure2013} and the assumption that $\tilde{\cal N} \geq \nu n$, 
\begin{eqnarray*}
\E\left[\sum_{t=1}^T R_t\right]
&\geq& \frac12 \left(\sum_{i=1}^m C_1 + C_2(\nu n)\log(T\mu_i/2)\right)\\
&\geq& \frac12 \left(\sum_{i=1}^m C_1 + C_2(\nu n)\log\left(\frac{T}{2 \rho m}\right)\right)\\
&\geq& \frac12 \left(C_1 m + C'_2(\nu n)\sum_{i=1}^m(\log T - \log(2 \rho m))\right).
\end{eqnarray*}
Hence if $m = o(T)$, we have
\[
\E\left[\sum_{t=1}^T R_t\right] = \Omega(mn \log T).
\]
(Otherwise, we have $\E[\sum_{t=1}^T R_t] = \Omega(mn \log(T/m))$.)
\halmos\endproof

Now we show that our structure aware algorithms produce regret
sublinear in $mn$.

\proof{Proof of Theorem \ref{regbound1}}

We begin by bounding the probability that the set $S^\star(w, \hat \Theta^{(t)}_{i_t}; K)$
that is offered in an exploitation round is different from the optimal set
$S^\star(w, \Theta^\star_{i_t}; K)$.

Note that, for any $S$ and $j\in S$,
$$
0\leq\frac{\partial p_j(S;\theta)}{\partial\theta_j}=\frac{e^{\theta_j}\sum_{j'\in S,\,j'\neq j}e^{\theta_{j'}}}{\prns{\sum_{j'\in S}e^{\theta_{j'}}}^2}\leq\frac{\abs{S}-1}{\abs{S}^2}\leq\frac14,
$$
and that, moreover, for any $j'\in S,\,j'\neq j$,
$$
0\geq\frac{\partial p_j(S;\theta)}{\partial\theta_{j'}}=\frac{-e^{\theta_j+\theta_{j'}}}{\prns{\sum_{j''\in S}e^{\theta_{j''}}}^2}\geq\frac{-e^{\theta_j+\theta_{j'}}}{\prns{e^{\theta_j}+e^{\theta_{j'}}}^2}\geq-\frac14,
$$
whereas, for $j'\notin S$, clearly $\frac{\partial p_j(S;\theta)}{\partial\theta_{j'}}=0$.
Therefore, $\magd{\nabla_\theta p_j(S;\theta)}_2\leq\frac14\sqrt{\abs S}$ for any $\theta\in\R n$.
This means that for any $S$, $j\in S$, $\theta$, and $\theta'$,
$$
\abs{p_j(S;\theta)-p_j(S;\theta')}\leq\frac14\sqrt{\abs{S}}\magd{\theta-\theta'}_2.
$$
It follows that for any $S$ with $\abs{S}\leq K$, $w$, $\theta$, and $\theta'$,
the expected revenue loss associated with choosing set $S'$ instead of $S$
is bounded by $\magd{\theta-\theta'}_2$:
$$
\abs{F(S;w,\theta)-F(S;w,\theta')}\leq\sum_{j\in S}\abs{w_j}\abs{p_j(S;\theta)-p_j(S;\theta')}\leq \frac14\magd{w}_\infty K^{3/2}\magd{\theta-\theta'}_2.
$$
Therefore, letting $S\in S^\star(w,\theta;K)$ and $S'\in S^\star(w,\theta';K)$,
\begin{align*}
F(S';w,\theta)
&\geq F(S';w,\theta')-\frac14\magd{w}_\infty K^{3/2}\magd{\theta-\theta'}_2\\
&\geq F(S ;w,\theta')-\frac14\magd{w}_\infty K^{3/2}\magd{\theta-\theta'}_2\\
&\geq F(S ;w,\theta )-\frac12\magd{w}_\infty K^{3/2}\magd{\theta-\theta'}_2.
\end{align*}
It follows that if $\magd{\theta-\theta'}_2$ is small enough,
the optimal assortment for each is the same:
$$
\prns{\magd{\theta-\theta'}_2\leq2\delta(w,\theta;K)K^{-3/2}/\magd{w}_\infty}\implies \prns{S^\star(w,\theta';K)=S^\star(w,\theta ;K)}.
$$

Now consider the regret $R_t$ incurred at time $t$ under the policy
$\pi_\text{nuc-norm}(C,\,\lambda)$.
If we explored at time $t$, then $R_t \leq \omega$.
If we exploited at time $t$, then
\begin{align*}
\E R_t
&\leq \omega \Prb{{S^\star(w,\theta';K)\ne S^\star(w,\theta ;K)}}\\
&\leq \omega \Prb{{\magd{\theta_{i_t}-\theta'_{i_t}}_2\leq2\delta(w,\theta;K)K^{-3/2}/\magd{w}_\infty}} \\
&\leq \omega \Prb{{\magd{\Theta-\Theta'}_2\leq2\delta K^{-3/2}/\magd{W}_\infty}},
\end{align*}
using that $\magd{W_i}_\infty\leq\magd{W}_\infty$, $\magd{\widehat \Theta_i-\Theta^\star_i}_2\leq\magd{\widehat \Theta-\Theta^\star}_\text{F}$, $0<\delta\leq \delta(W_i,\Theta^\star_i;K)$, and that the per-step regret is at most $\magd{W}_\infty$.
Now at each time $t$, note that the number of random observations (made in exploration rounds) is $N_t \geq C \log(t)$.
Then apply Theorem \ref{mainthm} with $\tau=\log(t)/\log(m+n)$ and use $\alpha\geq\alpha/\sqrt{mn}$
to see that
$$\E R_t \leq \omega \Prb{{\magd{\Theta-\Theta'}_2\leq2\delta K^{-3/2}/\magd{W}_\infty}} \leq \frac {3\omega} t.$$

Call the set of times when we explored $T_\text{explore}$, and note $|T_\text{explore}| \leq C\log(T) + 1$.
Summing this expression over $t$, we see
\begin{align*}
\op{Regret}_{\Theta^\star}^{\pi_1(T_0,\,\lambda)}
& \leq \sum_{t \in T_\text{explore}} R_t + \sum_{t \not \in T_\text{explore}} R_t \\
& \leq \sum_{t \in T_\text{explore}} \omega + \sum_{t \not \in T_\text{explore}} \frac {3\omega} t \\
& \leq  \omega (Cr(m+n) \log(T) + 1) + 3 \omega \log(T) = O(r\max(m,n)\log(T))
\end{align*}
which gives the result.
\halmos\endproof

\proof{Proof of Theorem \ref{thm:regret-dyn}}
Note that if $j\in S$ then
$$
\abs{\frac{\partial F(S;W,\Theta,\mu)}{\partial\Theta_{ij}}}\leq \mu_i\sum_{j'\in S}\frac{w_{ij'}}{4}\leq\frac{K\rho\magd{W}_\infty}{4m},
$$
and if $j\notin S$ then $\frac{\partial F(S;W,\Theta,\mu)}{\partial\Theta_{ij}}=0$. Moreover,
$$
\abs{\frac{\partial F(S;W,\Theta,\mu)}{\partial\mu_{i}}}=\abs{\frac{\sum_{j\in S}e^{\Theta_{ij}}w_{ij}}{1+\sum_{j\in S}e^{\Theta_{ij}}}}\leq\magd{W}_\infty.
$$
It follows that for any $S$ with $\abs{S}\leq K$, $W$, $\Theta$, $\mu$, $\Theta'$, and $\mu'$,
\begin{align*}
\abs{F(S;W,\Theta,\mu)-F(S;W,\Theta',\mu')}&\leq \frac{K\rho\magd{W}_\infty}{4m}\sum_{i=1}^m\sum_{j\in S}\abs{\Theta_{ij}-\Theta'_{ij}}+\magd{W}_\infty\sum_{i=1}^m\abs{\mu-\mu'}
\\&\leq \frac14{K^{3/2}\rho\magd{W}_\infty}\magd{\Theta-\Theta'}_\text{F}+\magd{W}_\infty\magd{\mu-\mu'}_1.
\end{align*}
Therefore, since the per-step regret is at most $\magd{W}_\infty$, we have that
\begin{align*}
\op{Regret}_{\Theta^\star}^{\pi_2(T_0,\,\lambda)}&\leq \magd{W}_\infty T_0 + \magd{W}_\infty T\times \Prb{\frac14{K^{3/2}\rho\magd{W}_\infty}\magd{\widehat\Theta-\Theta^\star}_\text{F}+\magd{W}_\infty\magd{\hat\mu-\mu}_1>\delta}
\\&\leq \omega T_0 +\omega T\times\prns{\Prb{\magd{\Theta-\Theta'}_\text{F}>\frac{2\delta}{K^{3/2}\rho\omega}}+\Prb{\magd{\mu-\mu'}_1>\frac{\delta}{2\omega}}}.
\end{align*}

Applying Theorem \ref{mainthm} with $\tau=\log(T)/\log(m+n)$ we get that
$\Prb{\magd{\Theta-\Theta'}_\text{F}>\frac{4\beta\delta}{K^{3/2}\rho\omega}}\leq 3/T$ given our choice of $T_0$.

By Theorem \ref{muthm}, $\Prb{\magd{\mu-\hat\mu}_1>\eta}\leq e^{m-T_0\eta^2/64}$. Since $T_0\geq\frac{256 \omega(m+\log(T))}{\delta^2}$ we have $\Prb{\magd{\mu-\mu'}_1>\frac{\delta}{2\omega}}\leq 1/T$. Finally, $\alpha\geq\alpha/\sqrt{mn}$ yields the result.
\halmos\endproof
\eeditb

\end{document}